# MatSegNet: a New Boundary-aware Deep Learning Model for Accurate Carbide Precipitate Analysis in High-Strength Steels

Xiaohan Bie[a], Manoj Arthanari[a], Evelin Barbosa de Melo[a], Baihua Ren[a], Juancheng Li[b], Nicolas Brodusch[a], Stephen Yue [a], Salim Brahimi [a], Raynald Gauvin[a], Jun Song[a*]

[a] Department of Mining and Materials Engineering, McGill University, Montréal, Québec H3A OC5, Canada

[b] Department of Mechanical Engineering, McGill University, Montréal, Québec H3A OC5, Canada

## Abstract

Lower Bainite (LB) and Tempered Martensite (TM) are two common microstructures in modern high-strength steels. LB and TM can render similar mechanical properties for steels, yet LB is often considered superior to TM in resistance to hydrogen embrittlement. Such performance difference has conventionally been attributed to their distinction in certain microstructural features, particularly carbides. The present study developed, MatSegNet, a new contour-aware deep learning (DL) architecture. It is tailored for comprehensive segmentation and quantitative characterization of carbide precipitates with complex contours in high-strength steels, shown to outperform existing state-of-the-art DL architectures. Based on MatSegNet, a high-throughput DL pipeline has been established for precise comparative carbide analysis in LB and TM. The results showed that statistically the two microstructures exhibit similarity in key carbide characteristics with marginal difference, cautioning against the conventional use of carbide orientation as a reliable means to differentiate LB and TM in practice. Through MatSegNet, this work demonstrated the potential of DL to play a critical role in enabling accurate and quantitative microstructure characterization to facilitate development of structure-property relationships for accelerating materials innovation.

# 1. Introduction

High-strength steels (HSS), distinguished by their remarkable strength, e.g., often with yield strength surpassing 460 MPa [1-3], are structural materials seeing widespread usage in numerous applications such as aircraft landing gears, construction, and automotive manufacturing [4-9]. Not only are they used in critical components requiring superior strength or load-bearing capacity, but they also play a vital role in reducing the weight of components. This makes HSS of strategic importance in enhancing energy efficiency and reducing emissions and carbon footprint. Lower Bainite (LB) and Tempered Martensite (TM) [8, 10] are two microstructures commonly present in many modern high-performance steels with superior mechanical properties. Both LB and TM are produced from medium carbon alloy steels but following different heat treatment processes. LB is a microstructure formed by austempering [8], where the steel is first austenitized above 800°C, then rapidly quenched to temperatures slightly above the MS temperature, and is held at the same temperature for a specified duration [11]. TM is produced with the steel first austenitized at temperatures above 800°C, and then quenched to a temperature below the martensite start (MS) temperature, where austenite transforms into martensite, followed by tempering at elevated temperatures below the austenitizing temperature [11]. Despite the significant difference in heat treatments, LB and TM steels share similar physical attributes, including hardness, modulus, and yield strength [8, 12, 13]. They also exhibit great similarity in the microstructures, both encompassing laths or plates of ferrite with finely dispersed carbide precipitates [8, 14].

Despite their similarities in physical properties and microstructural characteristics, LB and TM have been found to exhibit notably different performance in certain applications. One particular case is that recent experimental works [15-19] indicated that steel with the LB microstructure shows superior performance compared to steel of similar strength but with the TM microstructure, in resistance to hydrogen embrittlement (HE), a phenomenon where a material suffers loss of ductility, toughness and premature failure when exposed to hydrogen [20, 21]. The difference in HE susceptibility between LB and TM is often attributed to their distinction in certain microstructural features. In particular, a few studies suggested that carbides in LB microstructures tend to exhibit different morphologies [17] and be more finely dispersed [18], which favor more irreversible traps for hydrogen, thus yielding better mitigation against HE. These studies underscore the necessity of accurate characterization of carbides in LB and TM. Conventionally, characterization of carbides in LB and TM microstructures relies on microscopy techniques, commonly Scanning Electron Microscopy (SEM) and Transmission Electron Microscopy (TEM). Previous research based on these techniques reported that carbides in TM display random orientations while carbides in LB tend to be well aligned, orienting predominantly an angle of roughly 60° relative to the "growth direction" of ferrite plates [22, 23]. However, such differentiation between the two microstructures is qualitative and highly localized, typically based on features in individual limited fields of view [24], thus lacking statistical significance, and it has been shown to be not always reliable as significant variation in carbide orientations is also observed in LB [8]. In general, existing studies of LB and TM microstructures utilizing the conventional methods lack systematic and quantitative characterizations of the morphologies, distributions and alignment of carbides. Moreover, those methods involve considerable time-consuming manual work on the intricate images, thus highly dependent on experiences and prone to human bias.

In recent years, deep learning (DL) for computer vision has emerged as a powerful tool to potentially revolutionize microstructural characterization of materials [25-33], demonstrating unparalleled capabilities and efficiency in various image processing tasks such as classifications [34, 35], and segmentations [36-38], among others [39-41], and in many surpassing human experts in accuracy at a fraction of the time and cost [31, 42-44]. While Convolutional Neural Networks (CNNs), such as U-Net [26, 38, 45, 46] have long served as the *de facto* standard DL approach for segmentation tasks, the field has evolved to incorporate more advanced architectures to address complex morphological challenges. These include Feature Pyramid Networks (FPN) [47, 48], designed for multi-scale feature extraction, and transformer-based models, e.g., SegFormer [49-51], which leverage self-attention mechanisms to capture global contextual information, just to name a few.

Despite the convincing success demonstrated by DL in general microstructure characterization and analysis [27, 52], applying those generic DL architectures to high-strength steels remains challenging due to the inherent noise in secondary electron (SE) imaging and intricate topological variability in many microstructural features. One such microstructural feature group to note is secondary phases such as carbide precipitates.

Generic DL models often struggle to resolve the precise boundaries of densely packed or fine-scale phases, leading to under-segmentation or boundary blurring [26].

Motivated to address the afore-mentioned challenges, the present study introduces a new DL architecture, the Materials Segmentation Network (MatSegNet). The model integrates a gated attention mechanism [35, 53] with a dual-head output to jointly learn segmentation and boundary features, enhancing overall geometric fidelity. Scanning electron microscopy (SEM) characterization combining energy-dispersive spectroscopy (EDS) were performed to obtain high-resolution carbide morphologies for a robust ground truth. Applying MatSegNet in segmenting carbide precipitates in both LB and TM microstructure, we demonstrated that it outperforms other leading state-of-the-art DL architectures, in both segmentation accuracy and efficiency. Enabled by the quantitative analysis through MatSegNet, key aspects of carbide characteristics in LB and TM microstructures have been evaluated and systematically compared, providing robust statistics for reliable assessment of the microstructural similarity and difference in LB and TM. The workflow empowered by MatSegNet establishes bridges between DL and metallurgical physics to enable the autonomous and precise quantification of morphological features in the microstructures of engineering materials such as HSS.

# 2. Methodology

## 2.1 Materials

The samples used for microstructural characterization were made of medium-carbon, low-alloy steel, with the chemical composition summarized in **Table 1**. The relevant information and properties of the samples examined are presented in **Table 1**, while the material's two sets of samples were provided with identical chemical compositions but subjected to distinct heat treatments to render the TM (tempering at 440°C) and LB (austempering at 325°C) microstructures respectively in the samples. As can be noted in **Table 1**, the tensile and yield strengths of TM samples are slightly higher than those of LB samples. Charpy impact tests were also performed, from which the impact energy which quantitatively reflects the toughness of the steel [54], was measured, with the values listed in **Table 2**. We can see that the TM samples exhibit a slighter higher impact energy level, indicative of higher toughness.

**Table 1.** Chemical composition (in wt%) of the medium-carbon, low-alloy steel samples.

| Element | C | Mn | P | S | Si | Al | Ni | Cr | Cu | Mo | Nb | Ti | N | B |
|---|---|---|---|---|---|---|---|---|---|---|---|---|---|---|
| wt% | 0.44 | 0.94 | 0.005 | 0.005 | 0.17 | 0.025 | 0.16 | 1.48 | 0.04 | 0.19 | 0.05 | 0.01 | 0.005 | $8 \times 10^{-4}$ |

**Table 2.** Heat treatment temperatures, microstructures and key mechanical properties, i.e., yield strength (YS), tensile strength (TS) and Charpy impact energy at room temperature, for the two sets of steel samples considered.

| Heat Treatment | Temperature | Microstructure | YS (MPa) | TS (MPa) | Impact Energy (J) |
|---|---|---|---|---|---|
| Austempered | 325°C | LB | 1227 | 1538 | 10.2 |
| Quenched & tempered | 440°C | TM | 1455 | 1553 | 11.2 |

## 2.2 Sample preparation

The heat-treated wire rods were sectioned into 5 mm thick specimens using a laboratory-scale precision abrasive cutter. The samples were mounted in phenolic resin and mechanically ground using standard metallographic procedures up to 1200 grit SiC paper. Subsequent polishing was performed using 3 μm and 1 μm diamond suspensions to obtain a mirror-like surface finish.

## 2.3 SEM imaging for microstructure analysis

For scanning electron microscopy (SEM) imaging used to train and evaluate the DL model, specimens prepared according to the procedure above were etched using a 2% picral solution for 2-5 seconds to enhance microstructural contrast. The etching duration was adjusted within this range to ensure consistent feature delineation across samples.

SEM imaging was conducted using an FEI Quanta 450 scanning electron microscope operated in secondary electron mode. Microstructural features were examined at multiple magnifications, and imaging conditions were kept constant to minimize variability in contrast and brightness. A total of 40 SEM images were acquired and used for DL model development and analysis of the steel microstructure.

## 2.4 SE and EDS analysis for phase identification

To unambiguously identify the phases responsible for contrast variations observed in the SEM images, complementary secondary electron (SE) and energy-dispersive spectroscopy (EDS) analyses were performed on unetched, mirror-polished specimens using a Hitachi SU-9000EA scanning electron microscope. In addition to the specimen preparation procedure described in Section 2.2, the samples were subjected to a final polishing step using a 0.05 μm colloidal silica suspension mixed with 10 vol.% hydrogen peroxide ($H_2O_2$) to minimize surface deformation and enhance phase contrast.

SE imaging was carried out using the in-lens upper detector to maximize surface sensitivity and contrast between the carbides and the matrix. EDS data were acquired using an Oxford Instruments Extreme windowless silicon drift detector (80 $mm^2$). Low accelerating voltages were employed to improve spatial resolution and reduce interaction volume, specifically 1.5 kV for LB and 2.5 kV for TM microstructures. The chemical composition and elemental distribution of the matrix and interlath precipitates were characterized, and elemental maps were obtained for C, Si, Mn, S, O, Ni, Mo, and Cr. Point EDS analyses were performed on selected interlath precipitates and the surrounding matrix identified in SE images. The acquired spectra were normalized to the Fe-Lα peak at 0.71 keV to enable direct comparison of relative peak intensities and to confirm the presence of cementite precipitates.

## 2.5 Mask creation

Training the DL model starts with the creation of SEM image/mask pairs. These masks, representing the ground truth, were generated before training (Step 1, **Figure 1b**) and reflect the manual classification of the steel matrix and precipitates by expert metallurgists.

For making the masks, an approach based on pixel intensity values was implemented, where every pixel in the SEM images, ranging from 0 (black) to 255 (white) in intensity, was analyzed. A series of benchmark intensity values were selected for the identification of precipitates. The benchmark intensity is used as a threshold to classify pixels as dark (for the intensity below the threshold) and light (for the intensity above the threshold) areas, indicative of the steel matrix and precipitate region respectively (see **Figure 1a**), thus pixelwise classification of a SEM image. An iterative process was carried out in collaboration with experienced metallurgists to identify the best benchmark intensity that most effectively distinguishes the precipitates from background matrix regions for each SEM image with manual adjustments applied to minimize noise and smooth out uneven edges. After generating the carbide masks, the contours of individual precipitates were extracted for further analysis.

**Figure 1a–c** shows the segmentation results for a full-scale SEM image of the LB microstructure at 20000× magnification, with the overlay in **Figure 1c** demonstrating accurate capture of precipitate regions which are highlighted in pink. A representative region of interest (ROI) is cropped and magnified in **Figure 1d–f** for a clearer view of segmentation details. Even small and irregularly shaped precipitates are accurately delineated in both the precipitate mask (**Figure 1e**) and the corresponding edge mask (**Figure 1f**).

By feeding SEM images together with their corresponding mask (i.e., SEM image/mask pairs) into the model, the DL algorithm then learns from these expert classifications, to enable the model to independently classify similar patterns in new images.

## 2.6 Data Augmentation and Preprocessing

Given that only a limited number of SEM images were obtained, direct usage of these images as they are would provide insufficient input data for training a machine learning model [55]. To address this limitation, data augmentation was performed to expand the data set to obtain enough images for the training and later validation tasks. It is worth noting that here the data augmentation process is possible because we have high-resolution partitionable SEM images.

The data augmentation process starts by scaling an original SEM image to 10000X magnification, followed by cropping and cutting it into images of dimensions of 512 pixels in height and 512 pixels in width (see **Figure 2c**). It is important to note that the new image acquired from cropping and cutting still contains a considerable number of carbide and steel matrix features, sufficient to ensure that the essential microstructural signature is preserved to allow later training of the model. The data were then divided into three sets, namely the training (70% of data), validation (15% of data) and test (15% of data) sets. The image data were then input into different DL models.

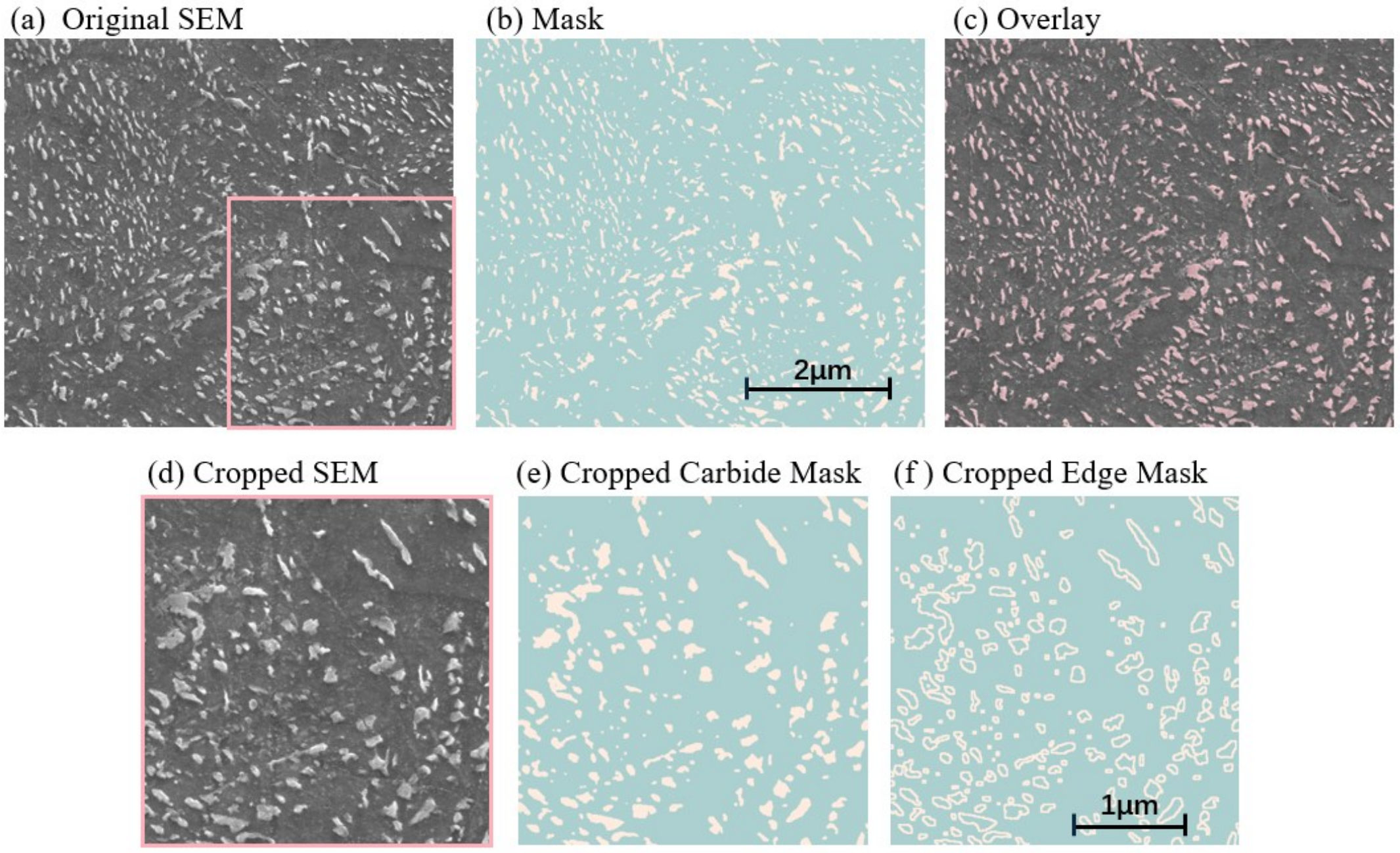

**Figure 1.** Visualization of SEM image segmentation and mask annotations. (a) A sample SEM micrograph of LB microstructure, at a magnification of 20000X , where the pink bounding box indicates the region of interest (ROI) cropped for panels (d–f). (b) The carbide mask created, where the cream and cyan regions represent precipitates and steel matrix respectively. (c) Overlay of the mask on the original SEM image, with precipitates highlighted in pink. (d) Cropped SEM micrograph of the ROI. (e) Corresponding cropped carbide mask. (f) Cropped edge mask delineating the carbide contours. Scale bars are 2 μm and 1 μm for (a–c) and (d–f) respectively.

## 2.7 Baseline DL models

For a comprehensive performance assessment of DL in microstructural characterization, three high-performance DL architectures, namely FPN, SegFormer and U-Net [36, 47, 50], were selected as the baseline models. They represent the current *state-of-the-art* models for semantic segmentation, regarded as the established benchmarks for tasks such as phase segmentation, microstructure characterization and fractographic analysis [26, 48, 56, 57], spanning the computational spectrum from local spatial biases to global contextual reasoning (**Table 3**).

The FPN architecture was implemented employing an EfficientNet backbone

(Supplementary **Figure S2**). This architecture is known for its robust capacity to resolve scale variance, enabling the simultaneous detection of both coarse and fine-grained microstructural constituents [47, 58]. The SegFormer is a leading transformer-based vision model [50]. Unlike traditional CNN, SegFormer utilizes a coordinate-free self-attention mechanism to capture long-range spatial correlations, a feature essential for analyzing complex, non-local microstructural patterns (Supplementary **Figure S3**). Meanwhile, the U-Net architecture considered here is a hybrid U-Net architecture integrated with a ResNet-34 encoder [35, 59]. Commonly regarded as the gold standard in scientific imaging, this U-Net architecture utilizes a symmetric encoder-decoder structure with skip-connections to ensure high-fidelity boundary and interface precision (Supplementary **Figure S4**). Detailed architectural configurations and implementation parameters of these baseline DL models are provided in the Supplementary Materials.

**Table 3.** Architectural characteristics and computational biases of the baseline DL models.

| **DL model** / **Property** | **FPN (Efficient Net)** | **SegFormer** | **U-Net (ResNet-34)** |
|---|---|---|---|
| Core Operator | Convolution (CNN) | Self-attention (Transformer) | Convolution (CNN) |
| Contextual Reach | Hierarchical multi-scale | Global microstructural context | Local neighborhood |
| Feature Integration | Top-down additive fusion | Unified MLP-based aggregation | Skip-connection (concatenation) |
| Morphological bias | Strong scale-invariance | Data-adaptive structural representation | Strong spatial locality |
| Scale adaptability | Multi-level hierarchical | Resolution-agnostic | Fixed or interpolated |
| Structural target | Multi-scale phase detection | Long-range spatial correlation | Boundary / interface precision |

## 2.8 New Materials Segmentation Network architecture

As demonstrated later in the results, though the three baseline models exhibit competitive performance, they do not fully satisfy the dual requirements of robust phase identification and precise boundary delineation. This limitation motivated us to develop a new DL architecture, namely the Materials Segmentation Network architecture (MatSegNet), details of which are illustrated in **Figure 2a**. The newly developed architecture, MatSegNet, begins with a 512×512×3 input image, and the encoder progressively reduces spatial dimensions (e.g., 512→256→128) while increasing feature depth (e.g., 3→32→64). Leveraging a ResNet-34 backbone [35], the encoder utilizes residual blocks to facilitate gradient flow and capture hierarchical features. These blocks employ 3×3 convolutions followed by batch normalization (BN) and ReLU activation to downscale the image, reducing its spatial dimensions and capturing abstract features [36]. At the bottleneck, the feature map is minimized (e.g., 16×16) with maximum depth (512 channels), bridging the encoder and decoder. The decoder then reconstructs the spatial resolution via 2×2 transposed convolutions, upscaling the feature maps back to the original sizes with the spatial resolution restored [36].

Crucially, attention-gated skip connections are integrated to refine the feature fusion between the encoder and decoder [53]. The attention coefficient, $\alpha_i$, which determines the importance of each spatial location, is computed as follows:

$$\alpha_i = \sigma\left(\psi^T\left(ReLU\left(W_x^T x_i + W_g^T g + b_g\right)\right) + b_\psi\right) \tag{1}$$

Where $x_i$ represents the high-resolution feature map from the encoder, preserving fine-grained spatial details and local structural contours while $g$ denotes the gating signal from the decoder, providing coarse-scale semantic information. The terms $W_x$ and $W_g$ are linear transformation matrices that project the encoder features and the gating signal into a common intermediate space. The parameter $\psi$ then aggregates the combined representation into a single scalar attention score. Following normalization via the sigmoid activation $\sigma$, the resulting coefficient $\alpha_i$ is applied to the original encoder features through element-wise multiplication, $\hat{x}_i = x_i \cdot \alpha_i$. This selective gating mechanism suppresses irrelevant background activations while emphasizing salient carbide structures, ensuring that only task-relevant information is propagated through the skip connections.

To improve computational efficiency, a tapered decoder was employed, progressively compressing feature channels (c.f. **Figure 2a**) as spatial resolution is gradually restored. By integrating transposed convolutions and bilinear interpolation for upsampling, this architecture facilitates robust high-level semantic integration in deeper layers while mitigating computational redundancy in high-resolution stages. The network terminates in a task-specific dual-head output architecture, i.e., the mask head performing pixel-wise semantic segmentation to distinguish precipitates from the steel matrix (with a total of 512×512 pixels), while the edge head simultaneously identifying precipitates' boundaries. Representative examples of the input image, the corresponding carbide mask, and the carbide boundary map are illustrated in **Figure 1d-f**. This joint optimization strategy allows the model to resolve complex carbide morphologies with high topological fidelity. Further details regarding the hyperparameters and layer configurations are provided in **Table SI** of the Supplementary Materials.

The performance of the proposed model was evaluated against the above three well-established segmentation models. A pipeline has been developed to streamline the automated analysis of carbide precipitates in HSS, schematically illustrated in **Figure 2b**. As seen from the pipeline, after obtaining the ground-truth masks, the next step is training the DL model.

The MatSegNet and three benchmark models were all trained over 35 epochs, with an epoch referring to a complete pass of the entire training dataset. Training over multiple epochs allows the model to iteratively learn from the data, adjusting its parameters to improve performance with each pass [60]. To mitigate overfitting, a two-stage transfer learning strategy was employed using a pre-trained encoder [61]. Initially, the encoder was frozen, and we trained only the decoder parameters for 30 epochs. Subsequently, the entire network was fine-tuned end-to-end with a reduced learning rate to adapt the encoder weights to our target domain, for an additional 5 epochs. This approach was further strengthened by applying weight decay ($L_2$ regularization [62]) and incorporating strategically placed dropout layers (see **Table SII** in Supplementary Materials for training hyperparameters). The optimal model was selected based on the peak *F1* score achieved on the validation set during training.

## 2.9 Performance Metrics and Convergence Analysis

The three baseline DL models and our newly developed MatSegNet were evaluated across two dimensions, i.e., *i*) predictive performance, measured by *accuracy*, *precision*, *recall*, *F1 score* and *loss* function, and *ii*) computational efficiency, quantified by parameters including floating-point operations (FLOPs), parameter count, inference latency per image, and frames per second (FPS).

For those metrics related to measuring the predictive performance, *accuracy*, *precision* and *recall* are defined as the follows:

$$Accuracy = \frac{\mathrm{TP} + \mathrm{TN}}{\mathrm{TP} + \mathrm{FP} + \mathrm{TN} + \mathrm{FN}} \tag{2}$$

$$Precision = \frac{\mathrm{TP}}{\mathrm{TP} + \mathrm{FP}} \tag{3}$$

$$Recall = \frac{\mathrm{TP}}{\mathrm{TP} + \mathrm{FN}} \tag{4}$$

Where TP (TN) denote true positives (true negatives), representing the precipitates (steel matrix) correctly identified by the model relative to the ground truth respectively, while FP (FN) indicate false positives (false negatives), representing regions labeled as the steel matrix (precipitates ) in the mask but predicted as the precipitates (steel matrix) by the model respectively.

The *F1 score* is the harmonic mean of *Precision* and *Recall*, serving as a balanced performance indicator, particularly valuable for imbalanced datasets:

$$F1\ score = \frac{2 * (Precision * Recall)}{Precision + Recall} \tag{5}$$

Since precipitates occupy only a small area fraction of the steel microstructure, simple pixel-level metrics (i.e., *accuracy*, *precision* and *recall*) could sometime be misleading due to the accuracy paradox in imbalanced datasets. The *F1 score*, being the harmonic mean of *precision* and *recall*, provides a more robust assessment of the model's ability to recognize the minority class (precipitates) without being inflated by the high background accuracy.

Meanwhile, the *loss* function quantifies the discrepancy between the predicted microstructural segmentation and ground-truth annotation, guiding the optimization of the network parameters during training [63]. Unlike those previously described performance measuring metrics, the *loss* function is not the same for different DL models. The FPN model was optimized using the binary cross-entropy (BCE) loss for its numerical stability and effectiveness in pixel-wise binary classification [60]. In contrast, U-Net and SegFormer were trained with the Dice Loss to mitigate the class imbalance inherent between the sparsely distributed precipitates and the iron background [64]. The Dice loss prioritizes regional overlap over pixel accuracy, which facilitates the precise delineation of fine structural features. While for MatSegNet, a hybrid loss function (**Eq. 6**) was adopted, which simultaneously optimizes carbide pixel classification and contour delineation. The first term is the BCE loss, with a weighting coefficient of $a$, primarily employed to enforce the pixel-level accuracy of the carbide mask [65]. The second term is the Dice loss, weighted at $b$, which specifically optimizes the prediction of carbide edges to mitigate the class imbalance inherent in edge detection [64]:

$$Loss = a \cdot \left( -\frac{1}{N}\sum_{i=1}^{N} y_i \cdot \log \hat{y}_i + (1 - y_i) \log (1 - \hat{y}_i) \right) + b \cdot (1 - \frac{2\sum_{i=1}^{N}(E_i \cdot \hat{E}_i)}{\sum_{i=1}^{N} E_i + \sum_{i=1}^{N} \hat{E}_i + \epsilon}) \quad (6)$$

Here, $a$ and $b$ are weighting coefficients. $y_i$ represents the carbide mask label, being 1 and 0 for carbide and steel background pixels respectively. In contrast, $\hat{y}_i$ denotes the predicted probability of a pixel point $i$ being a carbide pixel. Similarly, $E_i$ and $\hat{E}_i$ denote the ground-truth edge label and the predicted edge probability, respectively. A small constant $\epsilon$ is added to the denominator to prevent numerical instability. The coefficients $a$ and $b$ were empirically determined through tuning on the validation set; we assigned $a = 0.7$ and $b = 0.3$, which provided the most robust balance between pixel-level classification and boundary localization.

For the aspect of computational efficiency, we benchmarked the inference performance of different DL architectures on a standard consumer-grade GPU. The model's efficiency in terms of computational cost and inference speed was evaluated.

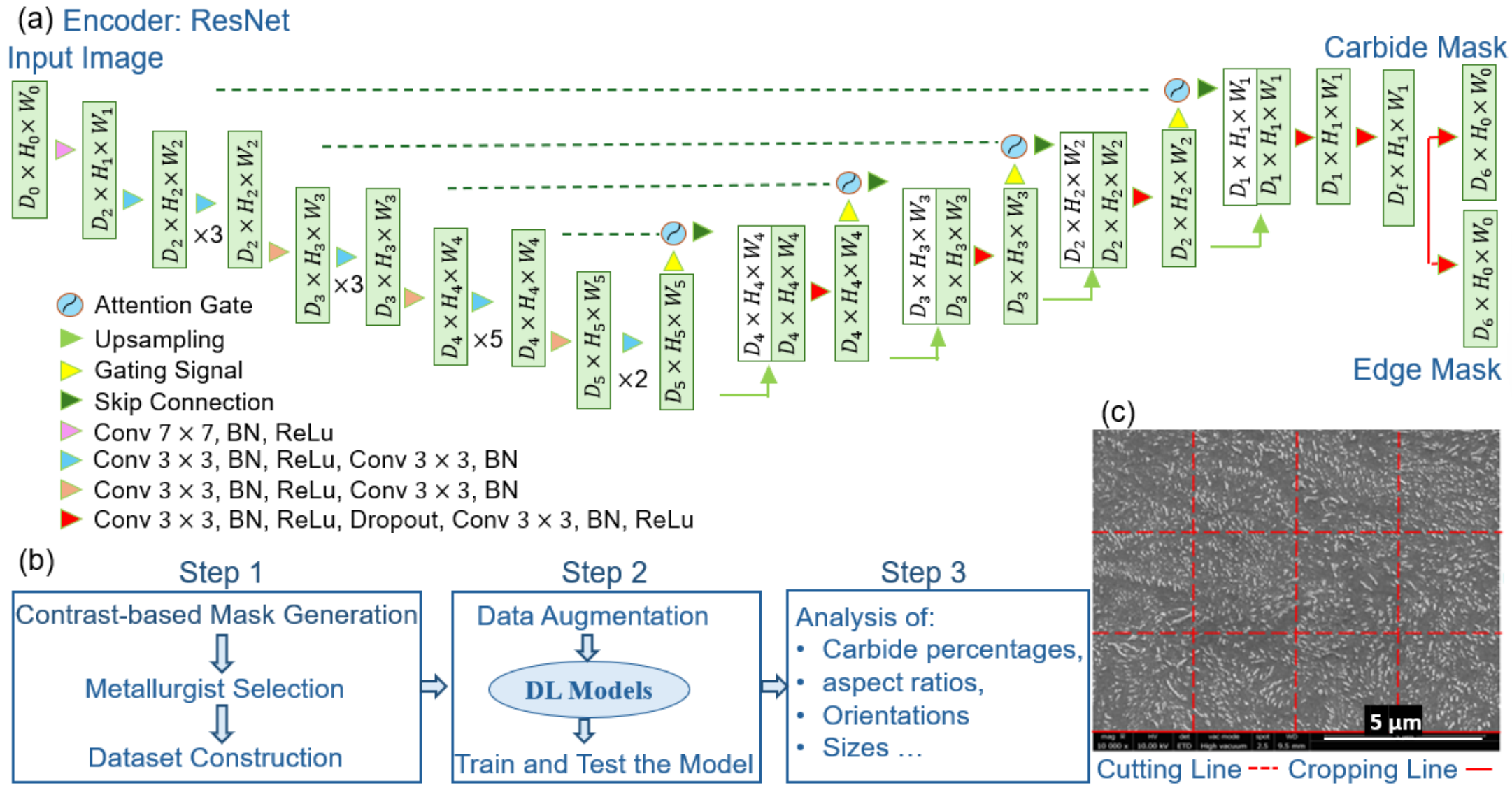

**Figure 2.** (a) The structure of the MatSegNet architecture, with each green box representing a multi-channel feature map. The notation D × H × W indicates the dimensions of the image, with H and W being spatial dimensions, and D denoting the number of channels (see values of different D, H, and W in **Table S1** in the Supplementary Material). White boxes denote copied feature maps, while the arrows indicate specific operations. (b) Steps for analyzing images. (c) Cropping of original SEM images.

# 3. Results and Discussion

## 3.1. Identification and Chemistry of Carbide Precipitates

SEM characterization confirms the LB and TM microstructures (**Figure S1**), both featuring fine carbide precipitates consistent with their respective heat-treatments. Detailed morphological verification and benchmarking against literature [66] are provided in Section S1 of the Supplementary Materials.

**Figure 3** shows the carbon map that correlate with the dark interlath particles observed in the SE image, indicating the presence of carbon enriched particles, presumable cementite precipitates. Maps of other elements commonly associated with precipitates, including O, S, Si, Mn, Mo, and Cr are homogeneous, showing no evidence of localized elemental enrichment (see **Figure S5**, Supplementary Materials). These observations indicate that compositional heterogeneity is largely confined to carbon, confirming that the observed precipitates are carbides, which are later determined to be cementite.

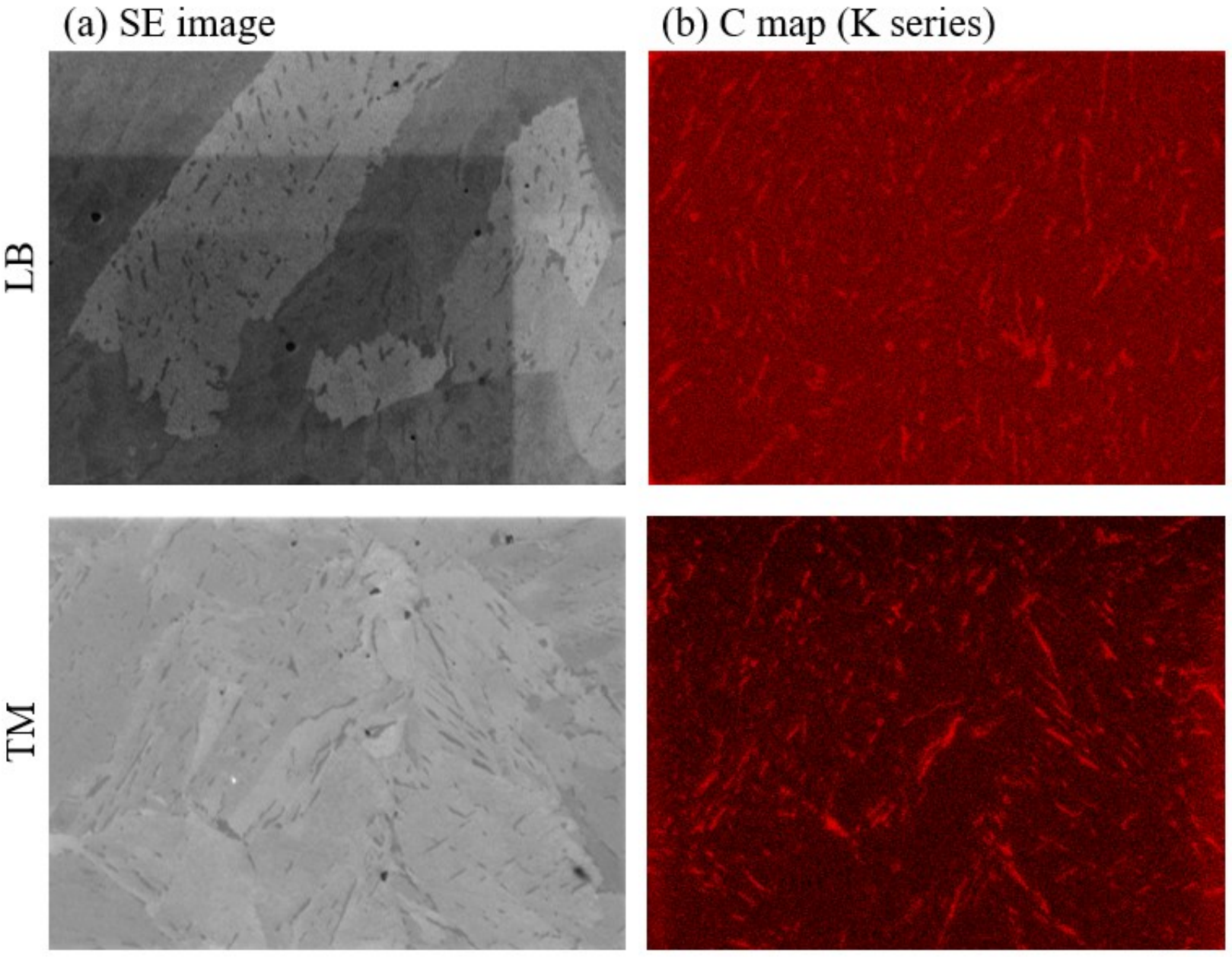

**Figure 3.** (a) SE images and corresponding low-voltage EDS elemental maps of (b) carbon for the LB and TM conditions.

Chemical compositions of the interlath precipitates were characterized via site-specific EDS point analysis, shown in **Figure 4**. To ensure quantitative comparability, all spectra were normalized to the Fe–Lα peak intensity (0.71 keV). In the LB sample, the C–Kα signal (0.27 keV) within the precipitates exhibited a normalized intensity of ~0.61, a three-fold increase over the matrix background (~0.22), confirming their identity as carbon-enriched cementite. A consistent trend was observed in the TM condition. Notably, the Mn–Lα profiles (~0.6 keV) remained invariant between the carbide and the adjacent matrix in both microstructures. This suggests a lack of substitutional solute partitioning, implying that carbide precipitation likely occurred under para-equilibrium or near-para-equilibrium conditions. Furthermore, high-fidelity point acquisitions revealed a distinct O–Kα peak (0.52 keV) localized at the LB cementite. This interfacial oxygen enrichment, which is undetectable in the coarser EDS maps, indicates preferential oxygen trapping or localized oxidation at the high-energy carbide/ferrite interfaces and is revealed here through the extended dwell times employed during point analysis.

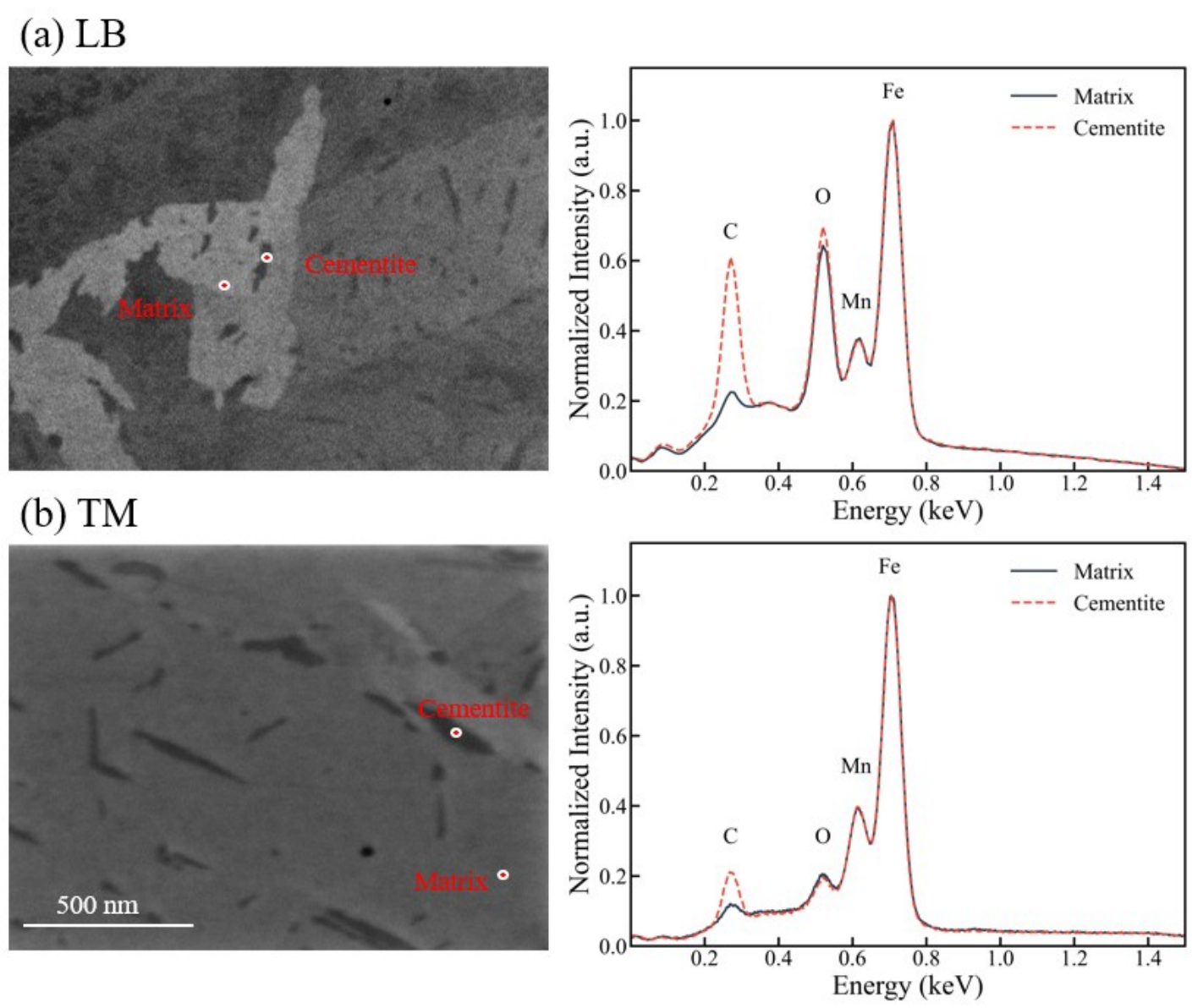

**Figure 4.** Representative SE images and corresponding normalized EDS spectra for (a) LB and (b) TM samples.

Following initial identification on unetched surfaces, etched samples were utilized to enhance carbide/matrix contrast for analysis. Due to their higher etching resistance, carbides protrude from the surface, yielding enhanced SE emission and a bright contrast relative to the dark matrix (**Figure 1a**) [66-68]. These light-gray protruding features correspond to cementite precipitates.

## 3.2 Microstructure Segmentation by DL Models

### 3.2.1. Overview of segmentation results

**Figures 5a** present representative cropped SEM images elucidating the LB and TM

microstructures respectively (see sample original, uncropped SEM images provided in the Supplementary Materials). Both microstructures can be seen to feature finely dispersed carbide precipitates, highlighted by the protruded regions. Visual comparison shows considerable resemblance between these two microstructures, making their differentiation challenging to naked eyes of non-experts. Utilizing the proposed pipeline (**Figure 2b**), the carbides and steel matrix in the microstructures were classified and analyzed by different DL models, with the results shown in **Figure 5a**.

To better visualize and compare the performance of different DL models, images with overlays of the results were presented in **Figure 5a**, highlighting false positives, false negatives, true positives and true negatives, which, as shown in **Eqs (2)-(5)**, are directly related to the several performance metrics. As seen from those overlay images, both LB and TM microstructures contain numerous needle-shaped precipitates. Though all DL models demonstrated good performance in identifying carbides, achieving accuracy values exceeding 98% (see **Table 3**), MatSegNet is seen to outperform the baseline models in producing the most accurate segmentation with the lowest false positives and false negatives. The enhanced accuracy from MatSegNet is found to come from it effectively suppressing the noises at the carbide/steel matrix interfaces, where the false positives and negatives concentrate, as illustrated in the high-resolution Region of Interest (ROI) zoom-in, shown in **Figure 5b**. This demonstrates the superior capability of MatSegNet in handling intricate contours over the baseline DL models.

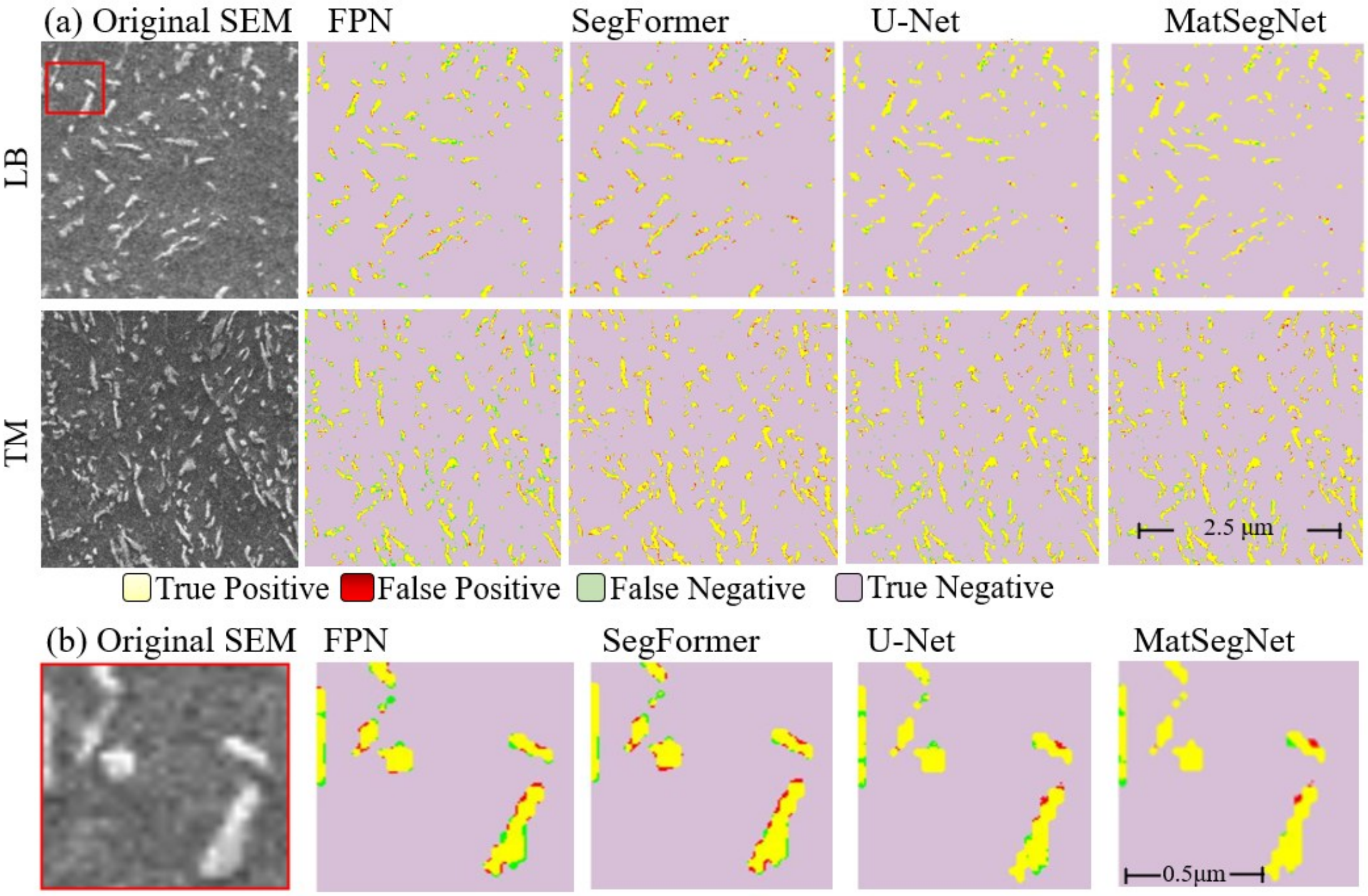

**Figure 5**: Comparison of different models for carbide segmentation in LB and TM samples. (a) Representative segmentation results (yellow: carbides; pink: iron matrix; red: iron matrix misclassified as carbides; green: carbides misclassified as iron matrix); the scale bar (top) represents 2.5 μm. (b) Magnified Regions of Interest (ROIs) corresponding to the red boxes in (a); the scale bar (bottom) represents 0.5 μm.

### 3.2.2. Segmentation performance comparison

Further to the results in **Figure 5** allowing visual comparison of the performance from different DL models, quantitative performance comparison has been conducted in terms of the metrics including *accuracy*, *recall*, *precision* and *F1 score*. MatSegNet achieved the highest performance across all metrics, followed by U-Net and then SegFormer, while FPN demonstrated the lowest performance.

**Table 3.** Evaluation of FPN, SegFormer, U-Net and MatSegNet were performed using metrics including *accuracy*, *recall*, *precision*, and *F1 score* (**Eqs. 2-5**).

| Model Name | Accuracy | Recall | Precision | F1 score |
| --- | --- | --- | --- | --- |
| FPN | 0.977±0.001 | 0.843±0.017 | 0.885±0.003 | 0.864±0.007 |
| SegFormer | 0.983±0.000 | 0.892±0.006 | 0.905±0.007 | 0.898±0.000 |
| U-Net | 0.988±0.000 | 0.929±0.002 | 0.930±0.001 | 0.929±0.000 |
| MatSegNet | 0.989±0.000 | 0.932±0.002 | 0.936±0.002 | 0.934±0.001 |

It is also worth mentioning that the performance of different models was monitored throughout the training process, to evaluate training behavior and model generalization. **Figure 6** presented the evolution of *loss* and *F1 score* for different DL models over 35 training epochs. As mentioned earlier, the *F1 score* provides a more robust assessment in recognizing the minority class (carbides) without being inflated by the high background accuracy. Therefore, for simplicity, the *F1 score* was selected as a primary performance metric. Meanwhile, the loss function quantifies the discrepancy between model predictions and the ground truth, and minimizing this loss guides the model to better fit the training data. As seen from **Figure 6**, all models exhibited stable convergence of both *loss* and *F1 score* on the validation dataset, indicating successful training without substantial overfitting. For the final evaluation, model checkpoints corresponding to the maximum validation *F1 score* achieved during the training epochs were selected. This early-stopping-like strategy ensures that the final model reflects the best generalization capability rather than overfitting to the training data. The curves shown in **Figure 6** represent a single representative training run, while similar convergence behavior was observed across runs. The quantitative results reported in **Table 3** are averaged over multiple runs.

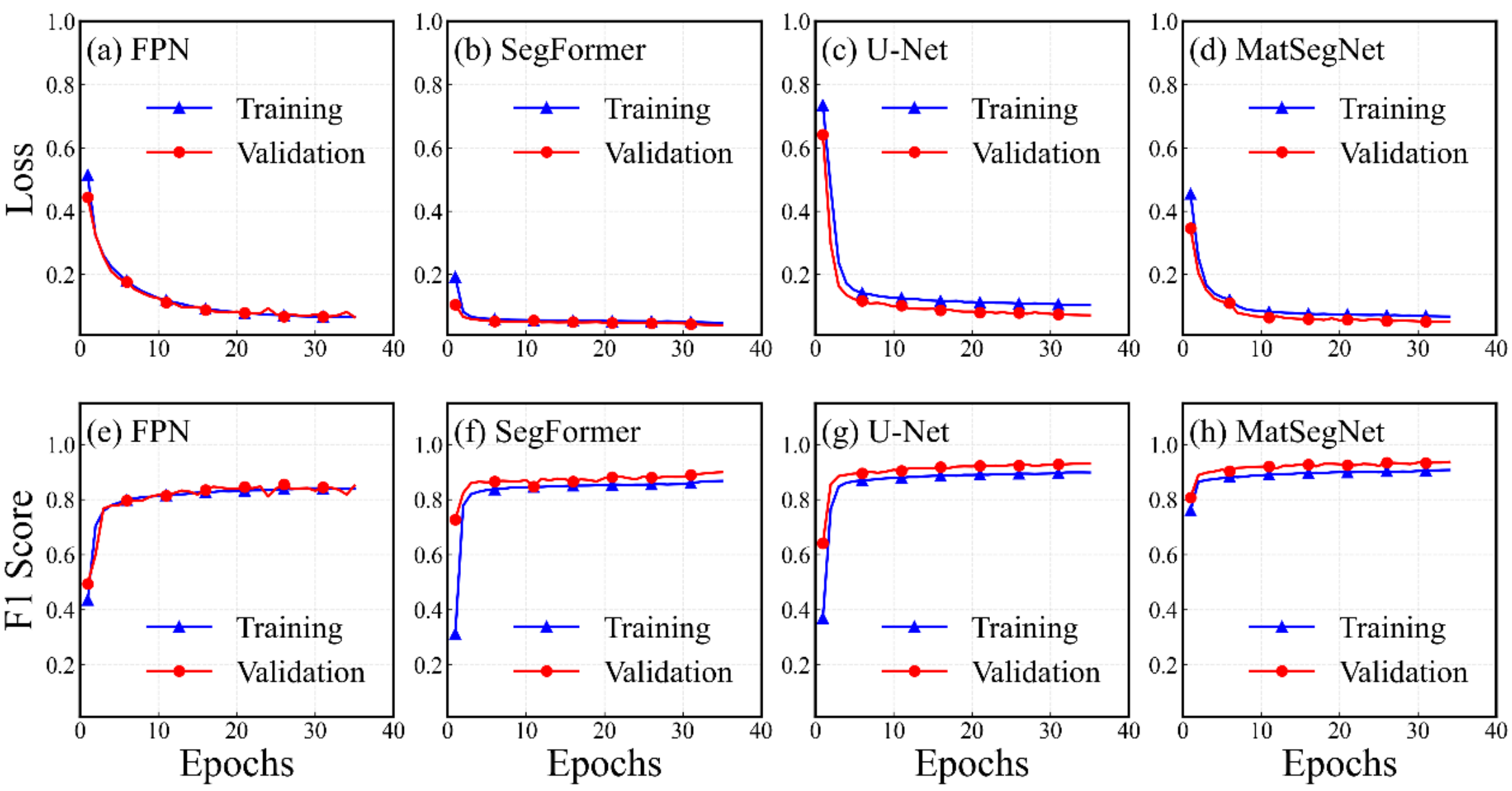

**Figure 6**. Evolution of training and validation, showing loss for (a) FPN, (b) SegFormer, (c) U-Net, and (d) MatSegNet, and F1 score for (e) FPN, (f) SegFormer, (g) U-Net, and (h) MatSegNet.

The metrics enlisted in **Table 3** as well as the loss function are based on pixel-wise predictions. In addition to these, it is also essential to evaluate the spatial fidelity of the segmented carbide regions, as it reflects the local agreement between predicted and ground-truth carbide distributions and is critical for assessing model performance across different regions of the image. For such purpose, we examined another parameter, i.e., the Intersection over Union (IoU), which is commonly used to evaluate how well a predicted region matches the ground truth [69-72]. IoU is defined as the ratio of the intersection to the union between the predicted and ground truth carbide regions (see **Figure 7b**):

$$IoU = \frac{area(C_p \cap C_{gt})}{area(C_p \cup C_{gt})} \tag{7}$$

Here, $C_p \cap C_{gt}$ represents the intersection of the predicted ($C_p$) and ground truth ($C_{gt}$) carbide regions, while $C_p \cup C_{gt}$ denotes their union. A high IoU, approaching 1, signifies a substantial overlap between the predicted shape and the ground truth, whereas a low IoU approaching 0 indicates a minimal intersection between the two. Given the large number of carbide regions in the image, we chose to compute IoU for carbides in each cropped image rather than for each carbide region. This patch-wise evaluation assesses the local spatial fidelity of the carbides and captures performance variability across different regions of the microstructure. To achieve this, the intersection (i.e., $C_p \cap C_{gt}$) was represented as the number of true positive pixels, while the union (i.e., $C_p \cup C_{gt}$) was taken as the addition of the numbers of true positive, false negative, and false positive pixels.

**Figure 7a** presents violin plots illustrating the distribution of IoU scores for the four DL models on the test dataset, revealing different distributional patterns. The shapes of the violins indicate that MatSegNet and U-Net are more robust as their predictions are consistently concentrated in the high-IoU range. In contrast, the elongated distribution of FPN, characterized by a downward tail, suggests higher sensitivity to local microstructural variation and less consistent segmentation quality. Notably, the IoU score distribution of MatSegNet is top-heavy, indicating that a large fraction of cropped images achieves superior segmentation accuracy. The superiority of MatSegNet over U-Net in terms of per-image IoU is supported by a Wilcoxon signed-rank test ($p < 0.0001$, denoted by the **** marker in **Figure 7a**), indicating that such per-image IoU difference is highly unlikely to arise from stochastic variation [73].

**Table 4** listed IoU statistics including mean, standard deviation, and quartiles. As expected, MatSegNet outperforms all baseline DL models, achieving the highest mean (0.880) and median (0.889) IoU values on the test set. The advantage of MatSegNet over others is also further supported by quartile analysis, with it achieving a notably higher third quartile (i.e., 0.940) than other models, indicating that its best-performing 25% of predictions are consistently more accurate. Importantly, all models show no signs of overfitting. The mean IoU values on the test set exceed those on the training set, which can be attributed to regularization and data augmentation during training (e.g., dropout and random rotations). Such heavy augmentation, i.e. rotation, increased the variability and complexity of the training samples, effectively making the training task more challenging than evaluating the original test images. Consequently, the models demonstrate superior performance on unseen data, indicating strong generalization capabilities.

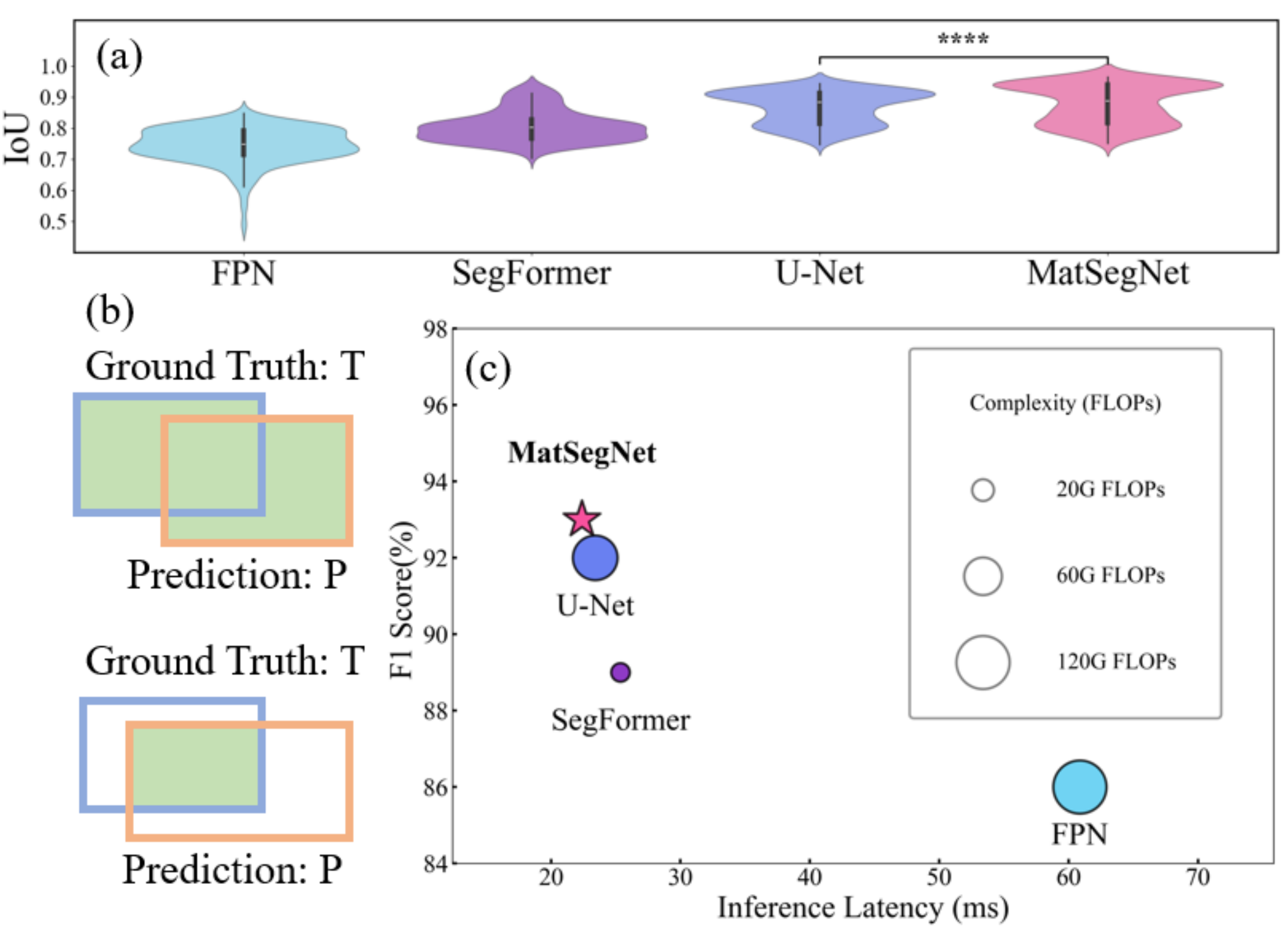

**Figure 7.** (a) Statistical comparison of IoU distributions on the test set. The significance marker **** represents $p$<0.0001 in Wilcoxon signed-rank test *[73]*. (b) Schematic diagram of the IoU

definition. (c) Speed-accuracy trade-off analysis benchmarked on an NVIDIA RTX 3050 Ti, where *x*-axis represents inference latency, and *y*-axis indicates the *F1 score*. Bubble and star sizes correspond to FLOPs.

**Table 4.** Statistical comparison of mean, standard deviation (Std Dev), Q1, median, Q3 values of IoU for different DL models.

| **Model** | Dataset | Count | Mean | Std Dev | Q1(25%) | Median | Q3(75%) |
|---|---|---|---|---|---|---|---|
| FPN | Training | 1094 | 0.7186 | 0.0535 | 0.6803 | 0.7228 | 0.7547 |
| | Validation | 235 | 0.7438 | 0.0564 | 0.7158 | 0.7440 | 0.7878 |
| | Test | 236 | 0.7456 | 0.0582 | 0.7152 | 0.7488 | 0.7920 |
| SegFormer | Training | 1094 | 0.7635 | 0.0527 | 0.7257 | 0.7578 | 0.8002 |
| | Validation | 235 | 0.8092 | 0.0502 | 0.7707 | 0.8034 | 0.8337 |
| | Test | 236 | 0.8076 | 0.0504 | 0.7675 | 0.8026 | 0.8278 |
| U-Net | Training | 1094 | 0.8226 | 0.0499 | 0.7878 | 0.8124 | 0.8484 |
| | Validation | 235 | 0.8688 | 0.0505 | 0.8224 | 0.8786 | 0.9132 |
| | Test | 236 | 0.8678 | 0.0528 | 0.8159 | 0.8846 | 0.9128 |
| MatSegNet | Training | 1094 | 0.8367 | 0.0526 | 0.8014 | 0.8340 | 0.8636 |
| | Validation | 235 | 0.8818 | 0.0606 | 0.8268 | 0.8884 | 0.9415 |
| | Test | 236 | 0.8804 | 0.0628 | 0.8170 | 0.8885 | 0.9395 |

### 3.2.3. Computational efficiency comparison

In addition to the performance of various DL models in characterizing steel microstructures, another important aspect critical to their deployment potential is the computational efficiency. Consequently, the efficiency was evaluated on a laptop-grade GPU (NVIDIA GeForce RTX 3050 Ti), as representative standard laboratory hardware. Four measurements, on different aspects of computational efficiency, were examined, including *Model Parameters*, *FLOPs*, *Inference Latency per image* (in milliseconds (ms)) and FPS, respectively indicating the memory footprint, theoretical computational cost and practical processing speed, i.e. time required to process a single image and the number of images a model can process per second. To ensure reproducibility, all benchmarks were standardized using a fixed input resolution of 512 × 512 pixels and a batch size of 1.

**Table 5** shows MatSegNet achieves the lowest latency (22.39 ms), outperforming the theoretically lighter SegFormer (14.78 GFLOPs). This discrepancy underscores that *FLOPs* fail to reflect actual latency, which is heavily influenced by hardware-level optimization [74]. By leveraging standard convolutions, MatSegNet achieves superior GPU utilization compared to transformer-based attention mechanisms [50, 75]. Additionally, MatSegNet reduces *FLOPs* by 25.7% compared to the strongest CNN baseline U-Net (from 84.73 to 62.93 GFLOPs), resulting in lower latency. This improvement stems from a tapered decoder design, i.e., U-Net keeps 64 channels in the final stages (c.f. **Figure S4**), whereas MatSegNet narrows the channels to 32 ($D_1$) and 16 ($D_f$) in the last upsampling blocks (c.f. **Figure 2**). Attention-gated skip connections [53] preserve spatial detail while avoiding redundant computation. Notably, MatSegNet retains this efficiency even with a dual-head output for simultaneous carbide and contour prediction.

**Table 5.** Computational efficiency comparison on 512×512 images (GPU: NVIDIA GeForce RTX 3050 Ti, batch size = 1). Units: *FLOPs* in GFLOPs (G = $10^9$), *parameters* in millions (M = $10^6$), *inference latency per image* in milliseconds (ms), and FPS.

| Model | *FLOPs* (G) | *Parameters* (M) | *Inference latency per image* (ms) (Mean ± Std ) | *FPS* |
|---|---|---|---|---|
| FPN | 120.00 | 22.20 | 60.85±7.26 | 16.43 |
| SegFormer | 14.78 | 3.75 | 25.37±5.33 | 39.41 |
| U-Net | 84.73 | 25.90 | 23.41±3.01 | 42.70 |
| MatSegNet | 62.93 | 24.44 | 22.39±2.71 | 44.64 |

The all-around edge of MatSegNet over the baseline DL models is further illustrated in the speed–accuracy analysis (**Figure 7c**), demonstrating it as an ideal DL architecture for both accurate and fast microstructural characterization.

### 3.2.4. Mechanistic analysis of performance advantages of MatSegNet

The superior performance of MatSegNet is fundamentally rooted in a task-specific design engineered to address the microstructural complexities of carbide-reinforced steels. Those baseline DL models, on the other hand, are limited in characterizing stochastic noise, complex boundaries, and the intricate morphologies commonly observed in SEM images.

The baseline DL models included two architectural paradigms, *i*) the CNN-based models, i.e., FPN and U-Net, and *ii*) a transformer-based model, i.e., SegFormer. Transformer architectures rely on tokenization, which partitions the input into patches, inherently reducing sub-patch spatial resolution. This process diminishes fine-grained spatial details during early feature encoding [50], resulting in smoothed or fragmented carbide boundaries and compromising the geometric precision required for rigorous microstructural analysis [76]. Meanwhile, for the CNN-based architectures, they also manifest limitations when handling complex phase boundaries. For example, while FPN-based models are effective in multi-scale feature aggregation, they lack the symmetric, high-capacity decoding structures necessary for precise pixel-level boundary refinement [77, 78]. The U-Net architecture, while well-suited for preserving high-resolution details through skip connections [79], is susceptible to propagating stochastic noise from SEM artifacts such as polishing scratches or etching-induced textures into the decoder, potentially leading to false-positive identification of carbide regions [59].

MatSegNet addresses the above challenges through architectural innovations, incorporating a gated attention mechanism [35, 53] alongside a dual-head output structure to simultaneously learn segmentation and boundary features, resulting in enhanced geometric fidelity. Attention gates act as adaptive semantic filters that suppress background noise while enhancing carbide-relevant features during upsampling [53]. The dual-head output structure decouples the learning objectives into complementary pathways, with a semantic head for phase-level segmentation and a boundary head for contour-level refinement. This design preserves topological integrity and improves segmentation along complex carbide boundaries. Additionally, a specialized loss strategy, i.e., weighed sum of BCE and Dice loss, was employed. The BCE loss is applied for the carbide mask for stable identification, whereas the Dice loss optimizes boundary prediction to address the extreme class imbalance inherent in the contour pixels. Such weighted loss strategy enables joint optimization of carbide segmentation and contour determination. Collectively, these designs allow MatSegNet to outperform both conventional CNN and Transformer models by effectively suppressing stochastic artifacts while preserving fine interfacial details, thus achieving high-resolution segmentation of complex multiphase microstructures.

With respect to computational efficiency, the transformer-based architectures like

SegFormer rely on self-attention mechanisms, which typically exhibit quadratic computational complexity with respect to the number of tokens ($O(N^2)$) [80], leading to increased inference latency at higher resolution, although recent designs mitigate this cost through efficient attention strategies [50, 81]. Meanwhile, CNN-based models primarily employ local convolution operations, which scale approximately linearly with the number of pixels (O($N$)) [82]. Consequently, at a moderate or high resolution, Transformer models are generally expected to exhibit higher inference latency compared to CNN-based models, which is the case for SegFormer compared to CNN-based U-Net and MatSegNet. However, FPN, also a CNN-based model, shows the highest inference latency. This exceptional high latency of FPN can be attributed to its multi-level upsampling and fusion introduce additional computational overhead, resulting in significantly longer inference time than others in practical deployment (c.f. **Table 5**). Meanwhile, among the CNN models, MatSegNet achieves the lowest theoretical FLOPs due to its tapered decoder design, and therefore the fastest inference speed. This combination of high-resolution accuracy and computational efficiency renders MatSegNet particularly suitable for real-world quantitative metallography applications.

## 3.3 Statistical Comparison of Carbides in LB and TM

Utilizing our newly developed DL architecture MatSegNet, which, as demonstrated above, achieves higher-fidelity segmentation results than the baseline models, a series of quantitative analyses of carbides in the LB and TM microstructures have been performed, detailed in the follows. Relevant analyses from other DL models were also performed (see details in Supplementary Material).

### 3.3.1 Carbide volume fraction

The accurate identification of carbides enables us to quantitatively analyze their features, including volume fraction, orientations, and morphology. Based on stereological principles [83, 84], the carbide volume fraction was determined directly from the area fraction measured on 2D micrographs. The carbide area fraction can be obtained by dividing the count of NN-predicted carbide pixels, by the total number of pixels within the images, as illustrated in **Figure 8a**. The volume fraction of carbides in each cropped image, identified by the image index, is plotted in **Figure 9a**, where the blue circles and purple triangles denote images from the LB and TM microstructures respectively. The carbide volume fraction distributions are shown in **Figures 9c** for LB and **9e** for TM. As can be noted from the two figures, carbides in both LB and TB exhibit a Gaussian-like distribution. We can also note from the results that there is a higher carbide volume fraction in the TM samples compared to the LB samples. The TM sample exhibited a mean volume fraction of 0.10, whereas the LB sample shows a mean of 0.08. The narrow 95% confidence intervals (**Figure 9a**) demonstrate the high precision of these results, confirming that our sampling methodology was sufficient to obtain stable and reliable averages for each condition.

### 3.3.2 Carbide orientation and alignment

In addition to the volume fractions, we also examined the orientation of carbides, as distinguishing carbide orientations is a crucial strategy for differentiating LB and TM [8]. The examination consists of three steps. First, each carbide was encapsulated in the smallest possible rectangle. To construct the enclosing box, we identified carbide contours using border-following algorithms [85] implemented by OpenCV [86]. The algorithm facilitated the extraction of a sequence of coordinates or chain codes from the border between the connected component of carbide pixels and the connected component of iron pixels. The bounding boxes were constructed based on the determined contours (see **Figure 8b**). Second, we determined the angle between the longer edge and the horizontal right direction. To ascertain the orientation is unique for each carbide, we calculated the angle to be within a range of -90° to 90° (see **Figure 8b**). Carbide regions (yellow areas in **Figure 8**) with an area smaller than 30 pixels, were classified as noises and thus excluded from the orientation analysis. Third, we gathered the orientation data and designed an alignment factor to characterize carbide orientations. The alignment factor, denoted as *k*, is calculated according to **Eq. 8** below, for each cropped SEM image:

$$k = \frac{\int_a^b f(x)dx}{\int_{-90^\circ}^{90^\circ} f(x)dx} \tag{8}$$

Here, $f(x)$ represents the number of carbides across different orientations $x$, spanning from -90° to 90°. We divided carbide orientations into 18 intervals (e.g. -90° to -80°, 80° to 90° for the first and last one) and found out the interval with the greatest number of carbides. The left and right bound of this interval is represented by $a$ and *b*. The larger the *k* value, the greater the alignment of the carbides. A higher *k* value indicates better alignment, and $k$ would approach 1 for near-perfect alignment. One important thing to note is that, given the scale of the cropped image, all the carbides therein primarily reside in a single lath.

The *k* values of the cropped images are displayed in **Figure 9b**. The outcomes reveal that the alignment factor *k* in LB and TM SEM images has averages of 0.24 and 0.17 respectively, with comparable standard deviation (STD) levels (being 0.07 and 0.04 in LB and TM respectively). The distributions of *k* values in LB and TM are shown in **Figures 9d** and **9f** respectively, both being Gaussian-like, with peak values of 0.23 and 0.15 respectively, while the distribution of *k* in LB displays a more pronounced tail in the high *k* range. Overall, the results indicate a more pronounced alignment of carbides in LB compared to TM. This is qualitatively in agreement with what's established in literature studies [8, 22, 23]. However, it is worth noting that no evidence of orientation-dominant carbide precipitation was identified in the present study. This observation is consistent with the established understanding that LB, similar to TM, does not necessarily exhibit a single crystallographic variant of precipitated carbide [8]. This suggests that using carbide orientation as the sole criterion for differentiating LB and TM is not always reliable.

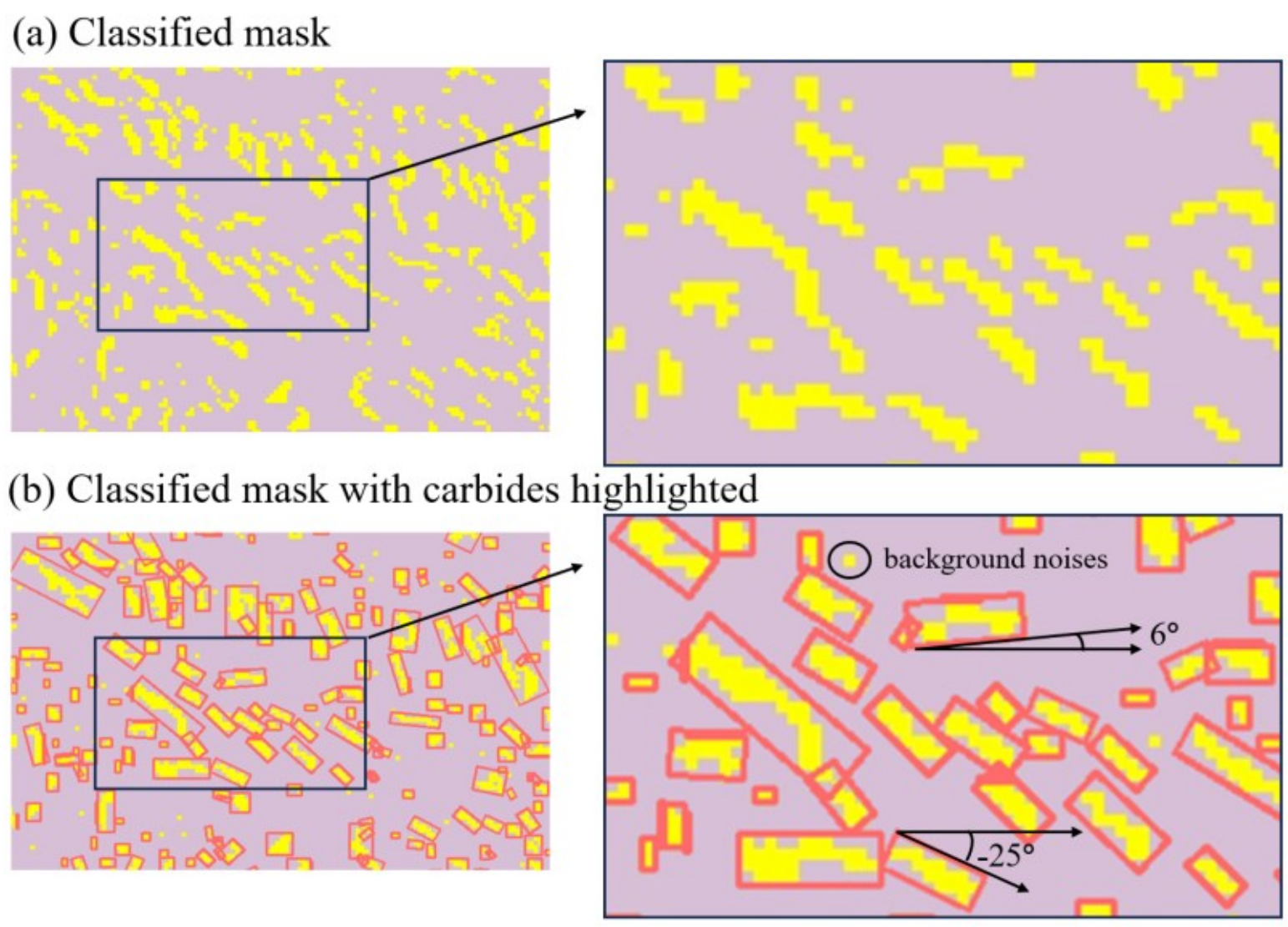

**Figure 8.** (a) DL classifications of carbides (yellow) and the iron matrix (pink), (b) Carbides are enclosed within defined boundaries (red rectangular boxes) for orientation characterization. The orientation angle is measured relative to the horizontal axis with a range from -90° to 90°.

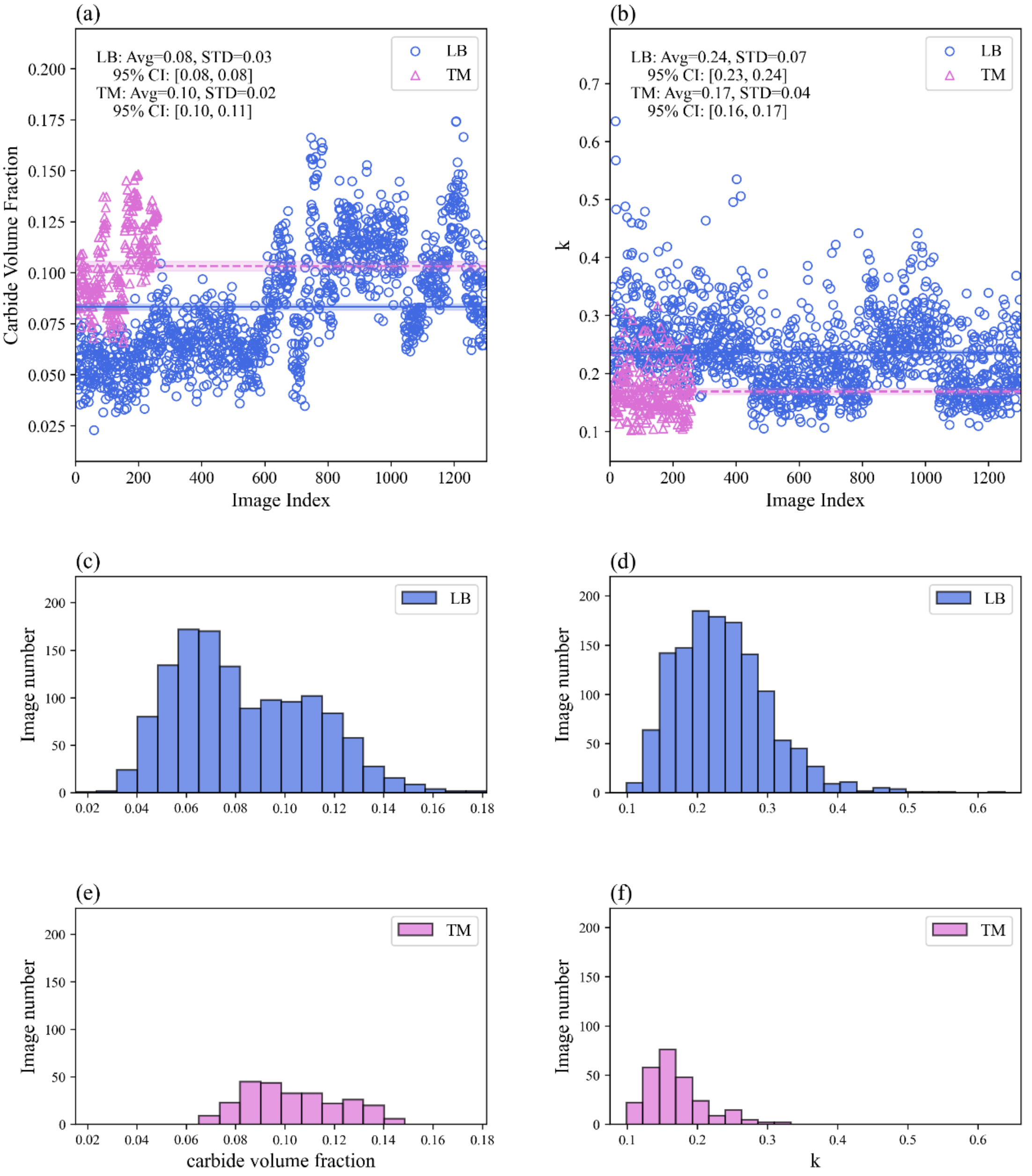

**Figure 9.** (a) Carbide volume fractions and (b) the alignment factor $k$ (see Eq. 8) plotted versus the image index. (c) and (f) represent carbide percentage distributions while (d) and (f) are distributions of $k$ in LB and TM respectively.

### 3.3.3 Carbide aspect ratio and size distribution

Furthermore, the aspect ratios and size distribution of carbides, which are also important metrics influencing the mechanical properties of HSS [87], have also been analyzed. Specifically, for the aspect ratio, we followed the same steps used in calculating the alignment factor as previously described, where the rectangular boxes enclosing carbides were determined. The value of the aspect ratio was then approximately calculated as the length (the longer edge) and width (the shorter edge) of the rectangular box for each carbide (see **Figure 8b**). The aspect ratio data are shown in **Figures 10a** and **10b** for the cases of LB and TM respectively.

At a high level, the aspect ratio distributions for the LB and TM samples appear qualitatively similar (**Figures 10c** and **10d**). Both are right-skewed and show a predominant population of carbides with aspect ratios between 1.00 and 2.43. However, a rigorous statistical analysis reveals that the distribution in the TM sample is shifted towards greater elongation. The mean aspect ratio for TM was 2.30 ± 1.00, notably higher than 2.14 ± 0.86 for LB. This difference is statistically significant, as confirmed by the non-overlapping 95% confidence intervals for their means ([2.29, 2.31] for TM vs. [2.13, 2.16] for LB). A similar significant shift is observed in the median values. Therefore, while both LB and TM show carbide populations with a similar morphological spread, the carbides in TM are consistently and measurably more elongated.

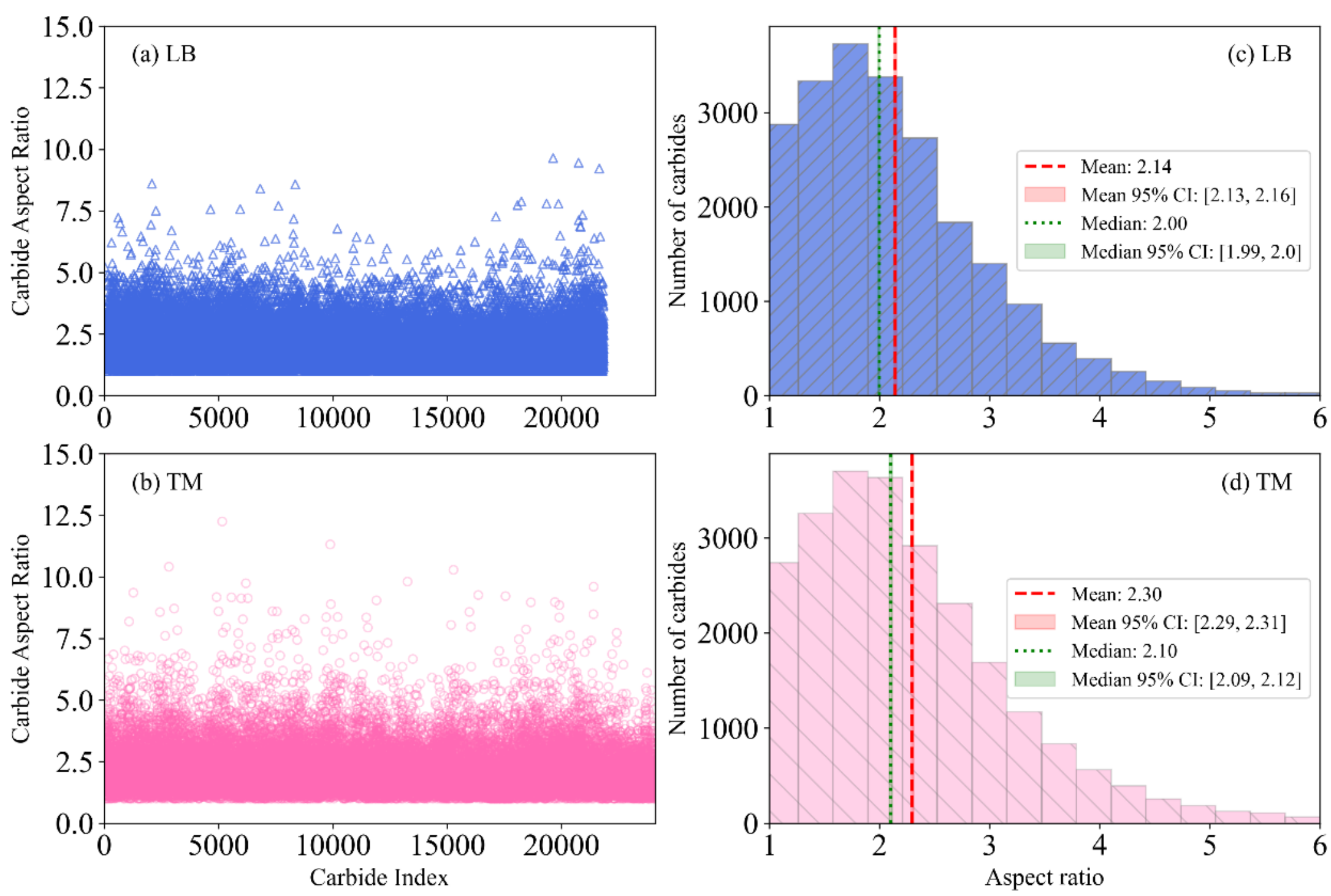

**Figure 10**. **(a-b)** The aspect ratios of carbides in LB (△) (std: 0.86 and mean: 2.14) and TM (○) (std: 1.00 and mean: 2.30). **(c-d)** Distribution plot of LB (/) and TM (\).

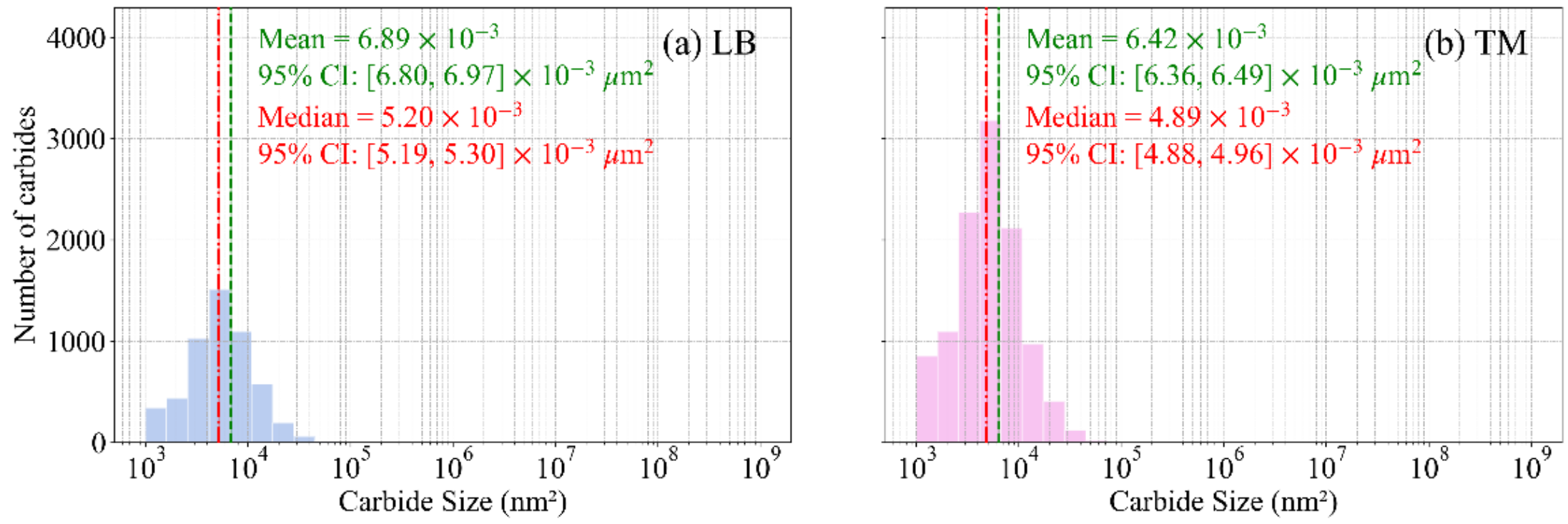

**Figure 11.** The size distributions of carbides (per $10^3$ $\mu m^2$) (a) in LB and (b) TM with their respective mean (green dash line) and median (red dash-dot line) size values indicated. (c) and (d) show the number of carbides with size greater than $10^5$ nm² in LB and TM respectively.

In addition to the aspect ratio, the sizes of carbides in LB and TM were quantitatively characterized. To ensure precise determination of the carbide size, the small, cropped images were at first amalgamated to the original un-cropped image size, mitigating any errors from the cropping of the carbides (see **Figure S7**). The sizes of carbides are determined by multiplying the number of pixels covering each carbide by the calculated pixel area, with additional details available in **Section S12** of Supplementary Material. Quantitative analyses of carbide size distributions for the LB and TM specimens are presented in **Figure 11**, showing that both microstructures exhibit characteristic right-skewed distributions, with the highest frequency of precipitates concentrated between $10^3$ and $10^4$ $nm^2$. Statistical evaluation reveals that the carbides in the LB specimen are coarser than those in the TM specimen, with carbides in LB exhibiting larger values in both mean and median carbide areas (see **Fig. 11**). Crucially, the 95% confidence intervals for the means (LB: $[6.80, 6.98]\times10^{-3}$ $\mu m^2$; TM: $[6.36, 6.49]\times10^{-3}$ $\mu m^2$) do not overlap, indicating that the observed smaller carbides in TM is statistically significant. This observation is further validated by the fact that the number of carbides in TM is greater than LB per 1000 $\mu m^2$.

These results are also consistent with the well-observed fact, highlighted by Bhadeshia [8], that TM generally exhibits a finer carbide dispersion than LB. Such difference in carbide dispersion between LB and TM may be attributed to higher thermodynamic driving force for carbide precipitation in the martensitic matrix owing to its greater carbon supersaturation [8], thus leading to a significantly increased nucleation rate during tempering [88].

# 4 Conclusion

In summary, the present study developed a new DL architecture, MatSegNet, tailored for the comprehensive segmentation and quantitative characterization of carbide precipitates with complex contours in high-strength steels. It has been demonstrated to achieve superior accuracy and efficiency, outperforming leading state-of-the-art DL architectures, represented by FPN, SegFormer, and U-Net. The advantages of MatSegNet over other DL models are rooted in a purpose-built dual-output structure for joint segmentation and boundary learning and task-oriented tapered decoder architecture, both of which are essential for accurately segmenting phases with complex contours.

Utilizing MatSegNet, quantitative characterization and statistical analysis were performed on various carbide characteristics in LB and TM microstructures. The TM microstructure was shown to exhibit a larger carbide volume fraction, *albeit* smaller carbide sizes, indicating more carbides precipitated during the tempering process for TM than the isothermal transformation in LB. This can be attributed to carbon supersaturation in the martensitic matrix increasing the thermodynamic driving force for cementite precipitation, thereby reducing the nucleation barrier to increase the nucleation rate. Meanwhile, carbide precipitates in TM were shown to be more elongated with higher aspect ratios. In addition, overall carbides were shown to be randomly oriented in LB and TM, with them being marginally better aligned in LB. Though the observation of better carbide alignment in LB than TM generally agrees with the conventional wisdom, it is in contrast with commonly used assertion of carbides in LB tend to exhibit a single variant with a well-defined angle relative to the "growth direction" of the ferrite plate, which is likely more applicable to local rather than global carbide arrangement. This also cautions against the use of carbide orientation as a reliable means to differentiate LB and TM in practices.

To the best of our knowledge, this study presents the first quantitative, statistical comparison of carbide characteristics in LB and TM. With the new DL architecture MatSegNet, the present study provides a workflow exceling at segmenting and characterizing phases with intricate interfacial geometries in the HSS microstructures. The success of MatSegNet lies in its emphasis on contour information. This contour-aware strategy, which jointly predicts phases and interfaces, is expected to generalize beyond carbide segmentation in HSS and extend to a broader range of microstructural systems and features in various materials. It provides a valuable tool for quick and accurate identification of critical descriptors in microstructures, thus facilitating development of structure-property relationships for accelerating materials innovation.

# Acknowledgements

The authors gratefully acknowledge the Arcelor Mittal Group (AMG) for supplying the materials used in this research. We also acknowledge financial support by Natural Sciences and Engineering Research Council of Canada (NSERC CRDPJ/543696-2019 and RGPIN-2023-03628), and sincerely thank the Digital Research Alliance of Canada for providing the high-performance computing resources. Special appreciation is given to Lucie Leclair for her invaluable contributions through insightful discussions during the study. Her expertise, guidance, and intellectual input significantly shaped the direction and enhanced the quality of this work. Special thanks are given to Prof. Xavier Feaugas for his valuable suggestions on statistical analysis of our results.

# Supplementary Materials

## S1. SEM Images of Lower Bainite (LB) and Tempered Martensite (TM) Samples after Nital Etching.

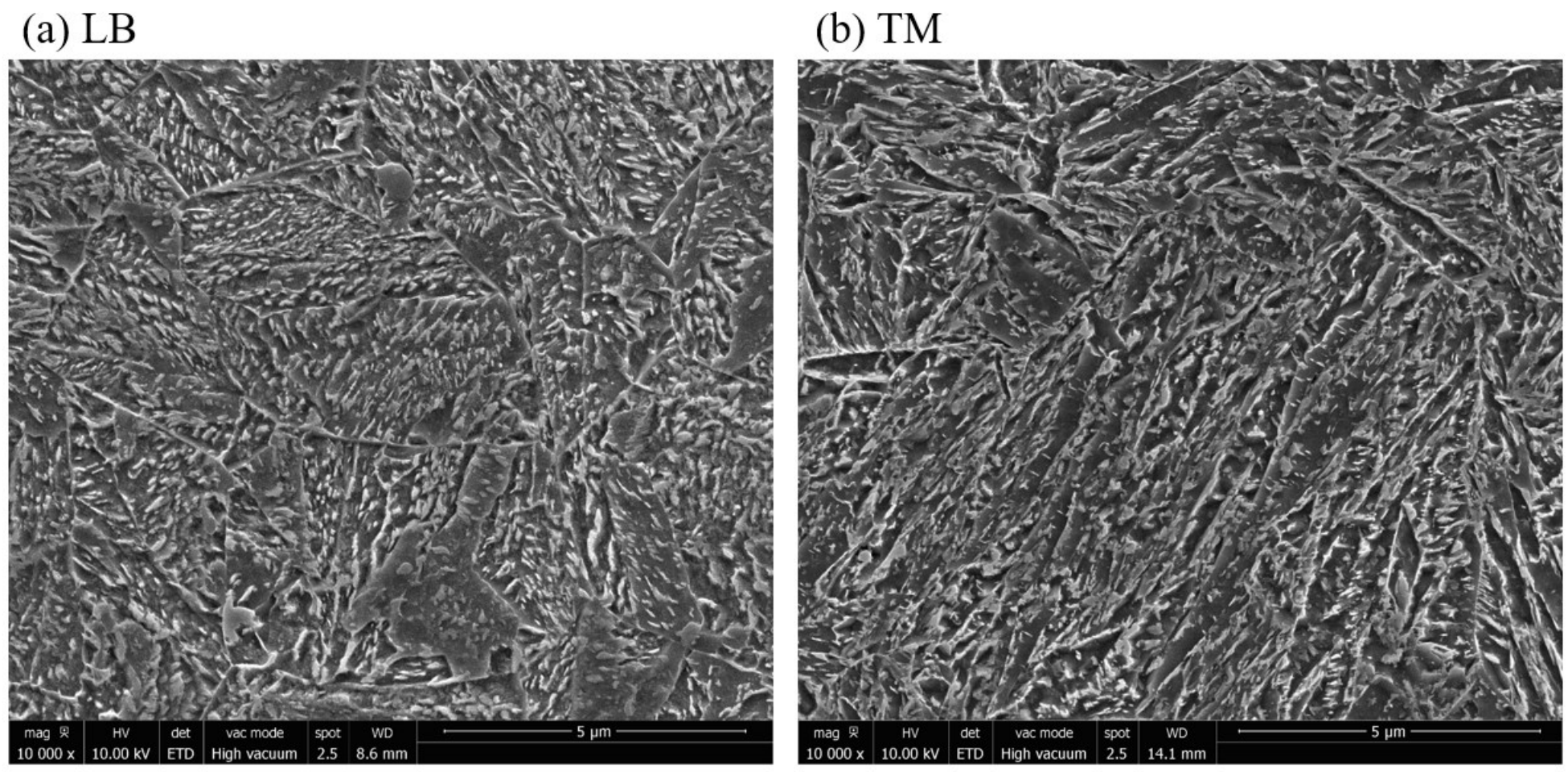

**Figure S1**. SEM micrographs of the (a) LB and (b) TM microstructures etched with 2% Nital.

**Figure S1** shows SEM micrographs of the (a) LB and (b) TM samples after etching with 2% Nital. The LB sample, austempered at 325 °C, exhibits a characteristic acicular morphology consisting of ferrite laths organized into parallel sheaves (**Figure S1a**). This morphology is consistent with the LB structures reported by Goulas et al. [1] for compositionally similar steels. As established via Transmission Electron Microscopy (TEM) and Atom Probe Tomography (APT) in literature [1], the defining feature of LB is the presence of fine, intralath carbides. The strong morphological similarity, combined with the specific austempering temperature used, confirms the phase to be LB, distinct from upper bainite or martensitic products.

Conversely, the TM sample (**Figure S1b**) exhibits the hierarchical lath morphology inherited from the parent martensite. Its defining feature is the precipitation of discrete cementite particles at both interlath boundaries and within laths. This multi-variant carbide precipitation habit, consistent with the prescribed tempering process, confirms the tempered martensitic microstructure.

## S2. Feature Pyramid Network (FPN)

The FPN architecture utilized in this study employs a pre-trained EfficientNet-B4 as its backbone [2]. The core objective of this design is to fuse rich semantic information from deep layers with fine-grained spatial details from shallow layers, thereby enabling accurate segmentation across multiple scales. The architecture comprises three distinct components:

1. *Bottom-up pathway (encoder)*: This pathway serves as the feature extractor. The input image propagates through successive layers to generate feature maps at decreasing spatial resolutions but increasing semantic depth (e.g., evolving from $3 \times 512 \times 512$ at the input to $1792 \times 16 \times 16$ in the deepest layer).
2. *Top-down pathway (decoder) and lateral connections*: The decoder upsamples high-level, semantically strong features to higher spatial resolutions. To preserve spatial detail, these upsampled features are merged with the corresponding feature maps from the bottom-up pathway via lateral connections. Specifically, the encoder features are projected using $1 \times 1$ convolutions to match the channel dimensions of the decoder. For instance, the P5 feature map is upsampled from $256 \times 16 \times 16$ to $256 \times 32 \times 32$, while the corresponding C4 map ($160 \times 32 \times 32$) is projected to 256 channels. These are then combined via element-wise addition.

3. *Segmentation head*: Finally, feature maps from all pyramid levels (P2–P5) are upsampled to a common resolution and concatenated along the channel dimension. A final classification layer produces the pixel-wise probability map for carbide segmentation.

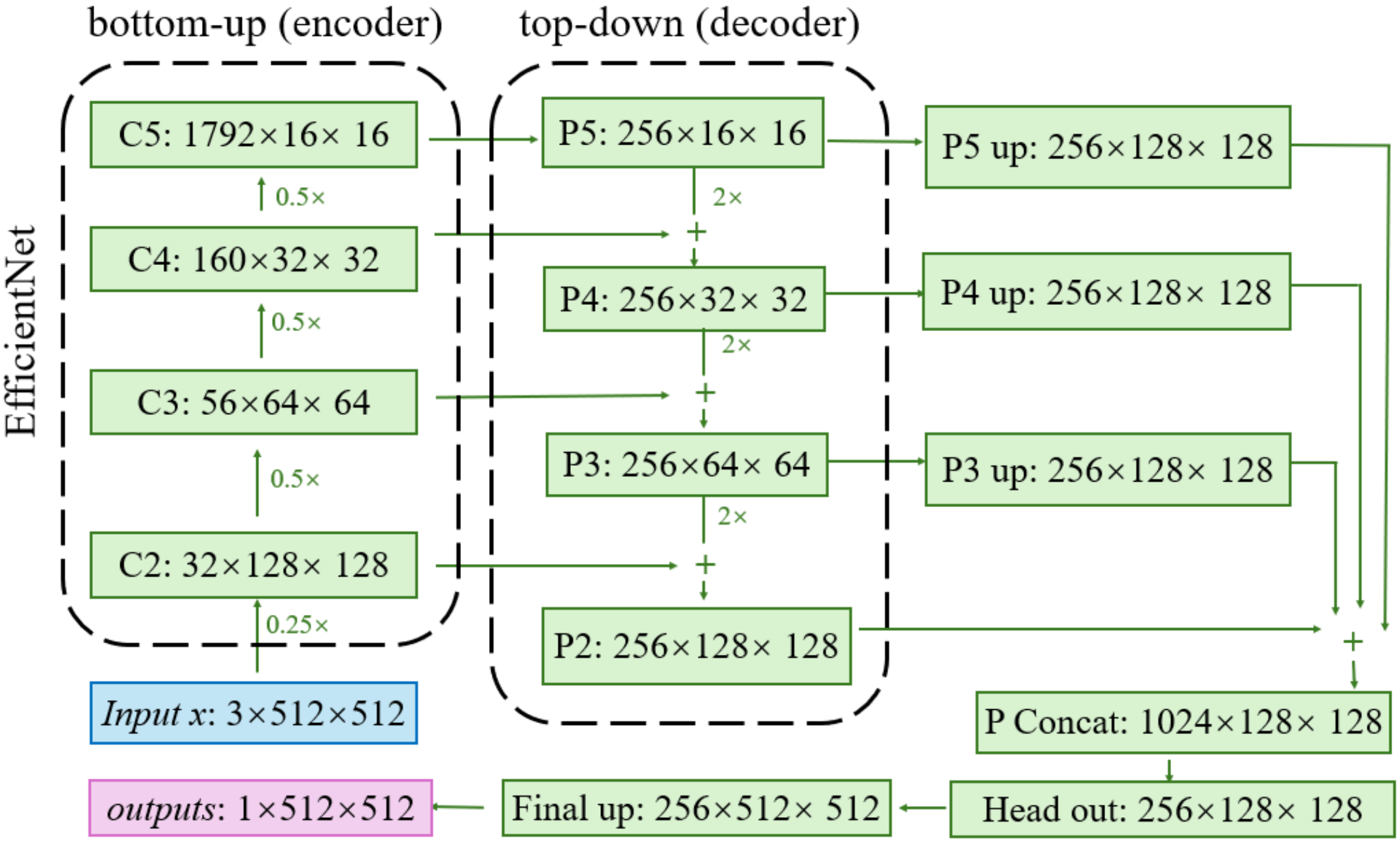

**Figure S2.** Schematic representation of the Feature Pyramid Network (FPN) architecture.

The model was trained using the binary cross-entropy (BCE) loss with logits [3], which integrates a sigmoid activation with standard binary cross-entropy. The loss is defined as:

$$Loss_{\mathrm{BCE}} = -\frac{1}{N}\sum_{n=1}^{N}\left[y_n \log \sigma(x_n) + (1 - y_n)\log\left(1 - \sigma(x_n)\right)\right] \quad \text{(S1)}$$

where $y_n$ denotes the carbide label (1 for carbide pixels, 0 otherwise), $x_n$ is the predicted logit for pixel $n$, and $\sigma(\cdot)$ is the sigmoid function.

## S3. Architecture of SegFormer

The architecture of the SegFormer model used in this study is illustrated in **Figure S3** [4]. SegFormer uses a hierarchical Vision Transformer (ViT) as an encoder to extract multi-scale features, enabling the simultaneous capture of precise local details and long-range contextual dependencies. Unlike standard Transformers, this architecture eliminates the need for positional embeddings, thereby offering greater flexibility for varying input resolutions. The decoder comprises a lightweight multi-layer perceptron (MLP) head that effectively integrates the encoder's multi-scale features to generate a pixel-wise segmentation map.

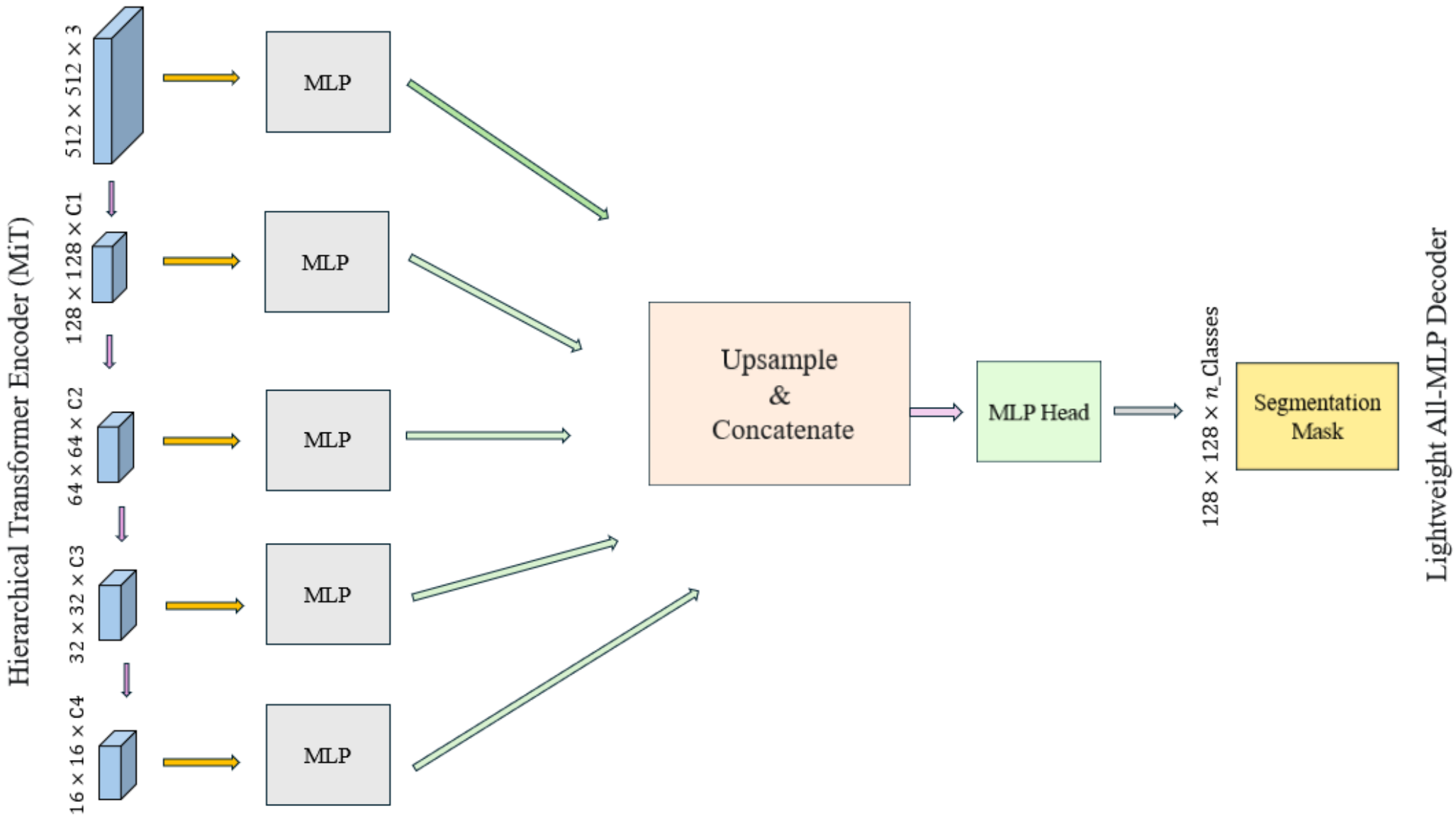

**Figure S3.** The architectural details of the SegFormer model.

The model was implemented using the SegformerForSemanticSegmentation class from the Hugging Face Transformers library [5]. The model was initialized with weights from the segformer-b0 backbone pre-trained on the ADE20K dataset. To adapt the architecture to our specific task, the final classification layer was modified to match the required number of output classes (binary: carbide vs. steel matrix), followed by fine-tuning on our prepared dataset.

The SegFormer model is trained using the Dice loss, which directly maximizes the overlap between predicted and ground-truth carbide regions [6]. The loss is formulated as:

$$Loss = 1 - \frac{2 \times \sum_{i=1}^{output\ size}(E_i \cdot \hat{E}_i)}{\sum_{i=1}^{output\ size} E_i + \sum_{i=1}^{output\ size} \hat{E}_i + \epsilon} \tag{S2}$$

where, $E_i$ denotes the ground truth label, taking a value of 1 for carbide pixels and 0 otherwise, and $\hat{E}_i$ represents the predicted probability that pixel $i$ belongs to the carbide class. $\epsilon$ is a small positive constant added to the denominator to prevent division-by-zero errors and ensure numerical stability. This loss function is particularly suitable for our imbalanced dataset, as carbide regions constitute only a small fraction of the entire image.

**S4. Architecture of U-Net**

The U-Net model employed in this study utilizes a ResNet-34 backbone as the encoder. The encoder progressively downsamples the input image (512×512×3) through a series of convolutional layers with a stride of 2, effectively reducing spatial resolution while increasing the feature channel dimensions. Four intermediate feature maps are extracted at varying scales, along with a bottleneck representation (16×16×512) that captures high-level semantic information.

In the decoder pathway, feature maps are progressively upsampled using 2 × 2 transposed convolutions. To recover fine spatial details lost during downsampling, each upsampled map is concatenated with the corresponding feature map from the encoder via skip connections. Following concatenation, the features are refined by two successive $3 \times 3$ convolutional layers, each integrated with batch normalization and ReLU activation.

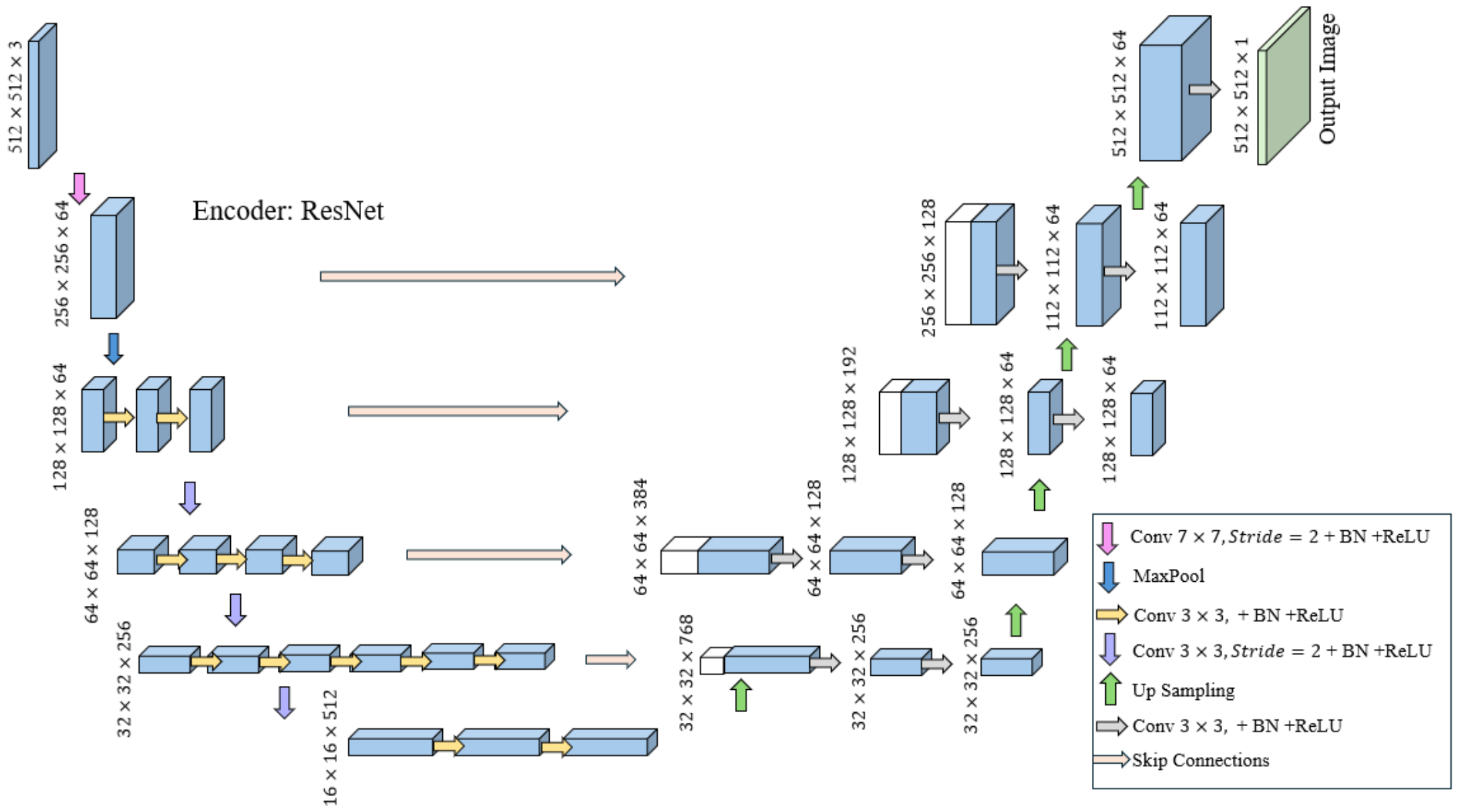

**Figure S4.** Schematic of the U-Net architecture.

Consistent with the SegFormer implementation, the U-Net model was trained using the Dice loss. The mathematical formulation and details are identical to those previously described (see **Eq. (S2)**).

## S5. Feature Map Dimensions in MatSegNet

The evolution of feature map dimensions throughout the MatSegNet architecture is summarized in **Table SI** and visualized in **Figure 2a**. This table tracks the changes in spatial resolution and channel depth, illustrating the progressive downsampling within the encoder and the subsequent upsampling across the decoder.

**Table SI.** Feature map dimensions ( $H \times W \times D$ ) at different stages of the MatSegNet architecture.

| $D_0$ | $D_1$ | $D_2$ | $D_3$ | $D_4$ | $D_5$ | $D_f$ | $D_6$ |
|---|---|---|---|---|---|---|---|
| 3 | 32 | 64 | 128 | 256 | 512 | 16 | 1 |

| $H_0 \times W_0$ | $H_1 \times W_1$ | $H_2 \times W_2$ | $H_3 \times W_3$ | $H_4 \times W_4$ | $H_5 \times W_5$ |
|---|---|---|---|---|---|
| $512 \times 512$ | $256 \times 256$ | $128 \times 128$ | $64 \times 64$ | $32 \times 32$ | $16 \times 16$ |

## S6. Training Protocols and Hyperparameters

The specific hyperparameters for each model and stage, including learning rates and epoch counts, are detailed in **Table SII**.

To mitigate overfitting, we applied consistent regularization techniques across all architectures, including $L_2$ regularization (weight decay) and tailored dropout strategies. For MatSegNet: we implemented a spatially adaptive dropout scheme within the decoder, with rates decreasing progressively from 0.4 in the deeper layers to 0.1 in the shallower layers. As for U-Net, spatial dropout with a decaying dropout rate is applied at the bottleneck and decoder layers to regularize deep features while preserving spatial details. And for SegFormer, dropout is implicitly applied through the Transformer encoder and the MLP decoder head. This includes attention dropout within self-attention layers and dropout in feed-forward and projection layers [4]. The dropout configuration follows the default settings of the pretrained SegFormer model and is not manually adjusted in this work.

**Table SII.** Hyperparameters of Different Models.

| Parameters | Unet | FPN | SegFormer | MatSegNet |
|---|---|---|---|---|
| Encoder | ResNet34 | EfficientNetB4 | MiT-B0 | ResNet34 |
| Batch Size | 16 | 6 | 16 | 8 |
| Epochs (Stage1)[a] | 30 | 30 | 30 | 30 |
| Epochs (Stage2)[b] | 5 | 5 | 5 | 5 |
| Learning Rate (Stage1) | 1.0e-4 | 3.0e-5 | 1.0e-3 | 1.0e-3 |
| Learning Rate (Stage2) | 1.0e-5 | 3.0e-7 | 1.0e-5 | 1.0e-5 |
| Weight Decay ($L_2$) | 1.0e-4 | 2.0e-4 | 0.01 | 1.0e-5 |
| Dropout | 0.4 → 0.1 | 0.3 | Default (built-in) | 0.4 → 0.1 |

$a$: Training with frozen encoder.

$b$: End-to-end fine-tuning.

## S7. EDS Maps for Other Elements

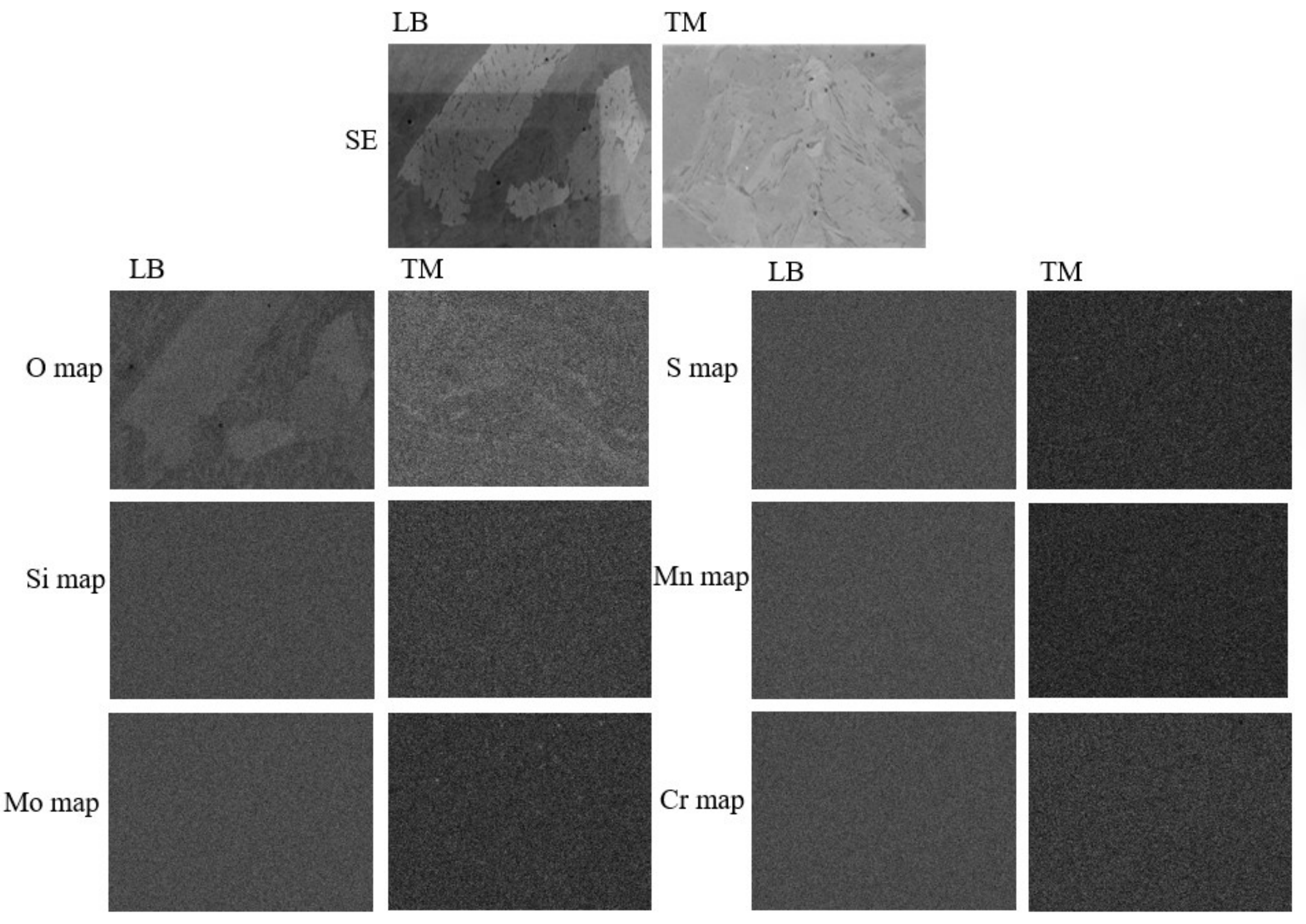

**Figure S5.** Complementary SE images and EDS elemental maps (O, S, Si, Mn, Mo, and Cr) for the LB and TM samples.

To evaluate the potential involvement of other alloying or impurity elements in the precipitation process, complementary EDS mapping was performed for O, S, Si, Mn, Mo, and Cr (**Figure S5**). The grayscale intensity across these maps remains statistically homogeneous and is consistent with the stochastic background noise. Crucially, no spatial correlation was observed between these elemental distributions and the black particles identified in the SE images. This lack of localized enrichment precludes the formation of oxides, sulfides, or complex silicides, confirming that the compositional heterogeneity is exclusively associated with carbide precipitates.

## S8. High Magnification Microstructures of LB and TM

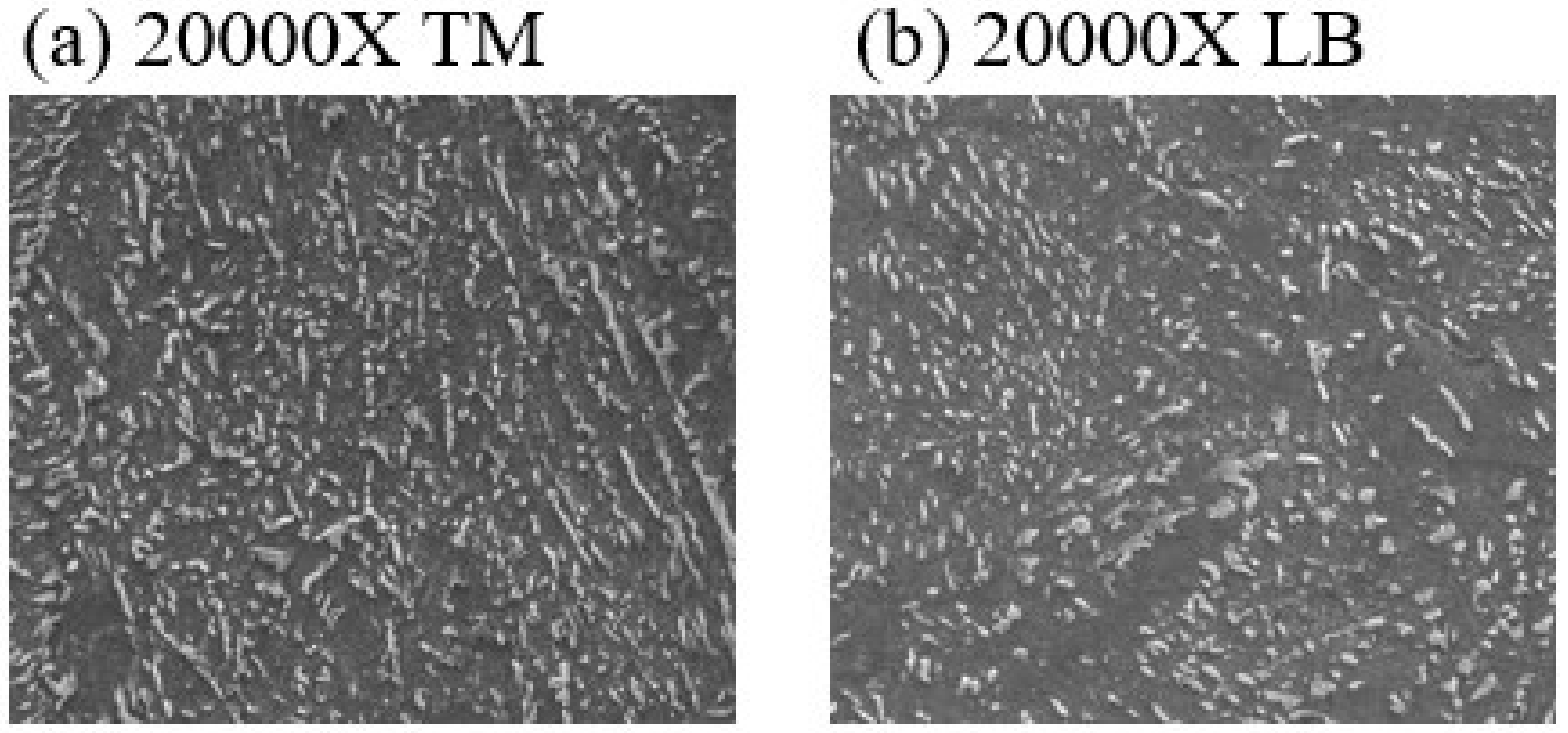

**Figure S6.** SEM micrographs at 20,000× magnification showing (a) TM and (b) LB.

**Figure S6** presents the microstructures of the (a) TM and (b) LB samples at 20,000× magnification. As observed, a significant number of fine precipitates are dispersed throughout the material, while the majority of the area corresponds to the steel matrix.

## S9. Cross-Model Validation of Carbide Volume Fraction and *k*

The statistical analysis of carbide volume fraction and $k$ from the predictions of FPN, SegFormer and U-Net is consolidated in **Table SIII**. Despite differences in network architectures, the three models exhibit highly consistent trends regarding the microstructural characteristics of the LB and TM samples.

As detailed in **Table SIII**, every model identifies a statistically significant difference in carbide volume fraction between the two datasets. The TM sample consistently shows a higher mean volume fraction (0.10) compared to the LB sample (0.08). The non-overlapping 95% CI across all models confirm the significance of this separation. Moreover, the distribution analysis reveals a structural difference discovered by all models: the LB dataset exhibits a bimodal distribution, indicating heterogeneity, whereas the TM dataset follows a more homogeneous, Gaussian-like distribution.

Conversely, the alignment factor $k$ shows the opposite trend. All models predict higher average $k$ values for the LB sample compared to the TM sample. While the absolute values vary slightly between models (e.g., SegFormer predicts slightly lower $k$ values than U-Net), the relative trend (LB > TM) and the distribution shapes (right-skewed for both datasets) remain robust and consistent.

The agreement across models reinforces the reliability of these microstructural observations, confirming that the reported differences between LB and TM samples are physical features of the material rather than artifacts of a specific segmentation algorithm.

**Table SIII.** Summary of statistical analysis for carbide volume fraction and alignment factor ($k$) across FPN, SegFormer and U-Net, comparing the LB and TM datasets.

| Model | Dataset | Carbide Volume Fraction [a] | Distribution Shape | Alignment Factor [a] ($k$) |
|---|---|---|---|---|
| FPN | LB | 0.08±0.03 [0.08, 0.08] | Bimodal | 0.19±0.05 [0.19,0.20] |
| FPN | TM | 0.10±0.02 [0.09, 0.10] | Gaussian-like | 0.17±0.04 [0.17,0.18] |
| SegFormer | LB | 0.08±0.03 [0.08, 0.08] | Bimodal | 0.19±0.05 [0.18,0.19] |
| SegFormer | TM | 0.10±0.02 [0.10,0.10] | Gaussian-like | 0.16±0.04 [0.16,0.17] |
| U-Net | LB | 0.08±0.03 [0.08,0.08] | Bimodal | 0.24±0.07 [0.24,0.24] |
| U-Net | TM | 0.10±0.02 [0.10,0.11] | Gaussian-like | 0.17±0.04 [0.16,0.17] |

$a$: Mean ± std [95% CI]

## S10. Cross-Model Validation of Carbide Aspect Ratio

The morphological characteristics of the carbides, specifically the aspect ratio, were analyzed to assess particles elongation. **Table SIV** summarizes the statistical comparison between the LB and TM datasets as predicted by FPN, SegFormer and U-Net.

A consistent trend is observed across both: carbides in the TM sample are statistically more elongated than those in the LB sample. For instance, SegFormer predicts a mean aspect ratio of 2.23 for TM compared to 2.10 for LB. A similar trend is seen with FPN(2.19 vs. 2.04) and U-Net (2.08 vs. 1.96).

The difference in elongation is statistically significant. As shown in **Table SIV**, the 95% CI for the mean aspect ratios of TM and LB do not overlap for either model (e.g., SegFormer: [2.22, 2.25] for TM vs. [2.09, 2.11] for LB). This significance is further corroborated by the median values, which are less sensitive to outliers. The median aspect ratios for TM (e.g., 2.04 for SegFormer) are consistently higher than those for LB (e.g., 1.94), with non-overlapping confidence intervals across all predictions.

While the absolute values vary slightly between architectures (with FPN and U-Net predicting slightly smaller aspect ratios), the relative morphological distinction is robust. The distributions for all models are right-skewed, indicating the presence of a tail of highly elongated carbides in both datasets, this feature is notably more pronounced in the TM samples.

**Table SIV**. Statistical summary of carbide Aspect Ratios predicted by FPN, SegFormer and U-Net. The table compares the Mean and Median values between LB and TM datasets.

| Model | Dataset | Mean Aspect Ratio[a] | Median Aspect Ratio[b] |
|---|---|---|---|
| FPN | LB | 2.04±0.81 [2.03,2.05] | 1.89 [1.88,1.90] |
| FPN | TM | 2.19±0.95 [2.18,2.20] | 2.00 [2.00,2.00] |
| SegFormer | LB | 2.10±0.83 [2.09,2.11] | 1.94 [1.93,1.96] |
| SegFormer | TM | 2.23±0.97 [2.22,2.25] | 2.04 [2.03,2.06] |
| U-Net | LB | 1.96±0.76 [1.95,1.97] | 1.82 [1.81,1.83] |
| U-Net | TM | 2.08±0.89 [2.07,2.09] | 1.91 [1.90,1.92] |

$a$: Mean ± Std [95% CI]
$b$: Median [95% CI]

## S11. Cross-Model Validation of Carbide Size Analysis

To characterize the particle size distributions, the carbide areas were analyzed across all model predictions. The statistical summary is presented in **Table SV**. A consistent statistical pattern is observed across all three evaluated models (FPN, SegFormer, and U-Net). Across all architectures, the LB samples consistently exhibit higher mean carbide sizes than the TM samples. For instance, in the FPN model, the mean area for LB is $6.96\times10^{-3}$ $\mu m^2$ compared to $6.00\times10^{-3}$ $\mu m^2$ for TM. All models confirm that the median carbide size in the LB sample (ranging from 5.35 to $5.62\times10^{-3}$ $\mu m^2$) is statistically larger than that in the TM sample (ranging from 4.70 to $5.21\times10^{-3}\mu m^2$). Most importantly, the non-overlapping 95% CI for both the mean and medians across all models confirm the statistical significance of this observation. These findings consistently demonstrate that the TM sample contains finer carbides compared to the LB sample, reflecting the enhanced nucleation and restricted growth kinetics characteristic of martensite tempering.

**Table SV.** Statistical comparison of carbide size (area) distributions predicted by FPN, SegFormer and U-Net.

| Model | Dataset | Mean Size ($\times10^{-3}$ $\mu m^2$) (Mean [95% CI]) | Median Size ($\times10^{-3}$ $\mu m^2$) (Median [95% CI]) |
|---|---|---|---|
| FPN | LB | 6.96 [6.89,7.04] | 5.41 [5.35, 5.50] |
| FPN | TM | 6.00 [5.94,6.06] | 4.70 [4.67, 4.74] |
| SegFormer | LB | 7.04 [6.96,7.13] | 5.35 [5.30, 5.41] |
| SegFormer | TM | 6.48 [6.41,6.56] | 4.99 [4.89, 4.99] |
| U-Net | LB | 7.36 [7.27,7.45] | 5.62 [5.52,5.65] |
| U-Net | TM | 6.91 [6.84,6.99] | 5.21 [5.20,5.30] |

## S12. Carbide Quantification in Reconstructed Images

**Figure S7a** displays the original uncropped SEM micrograph. To quantify the carbide area, the individual cropped classification results were recombined. The resulting composite mask is shown in **Figure S7b**, where pink regions correspond to the steel matrix and yellow regions indicate carbides. Carbide area calculations were performed using this reconstructed image.

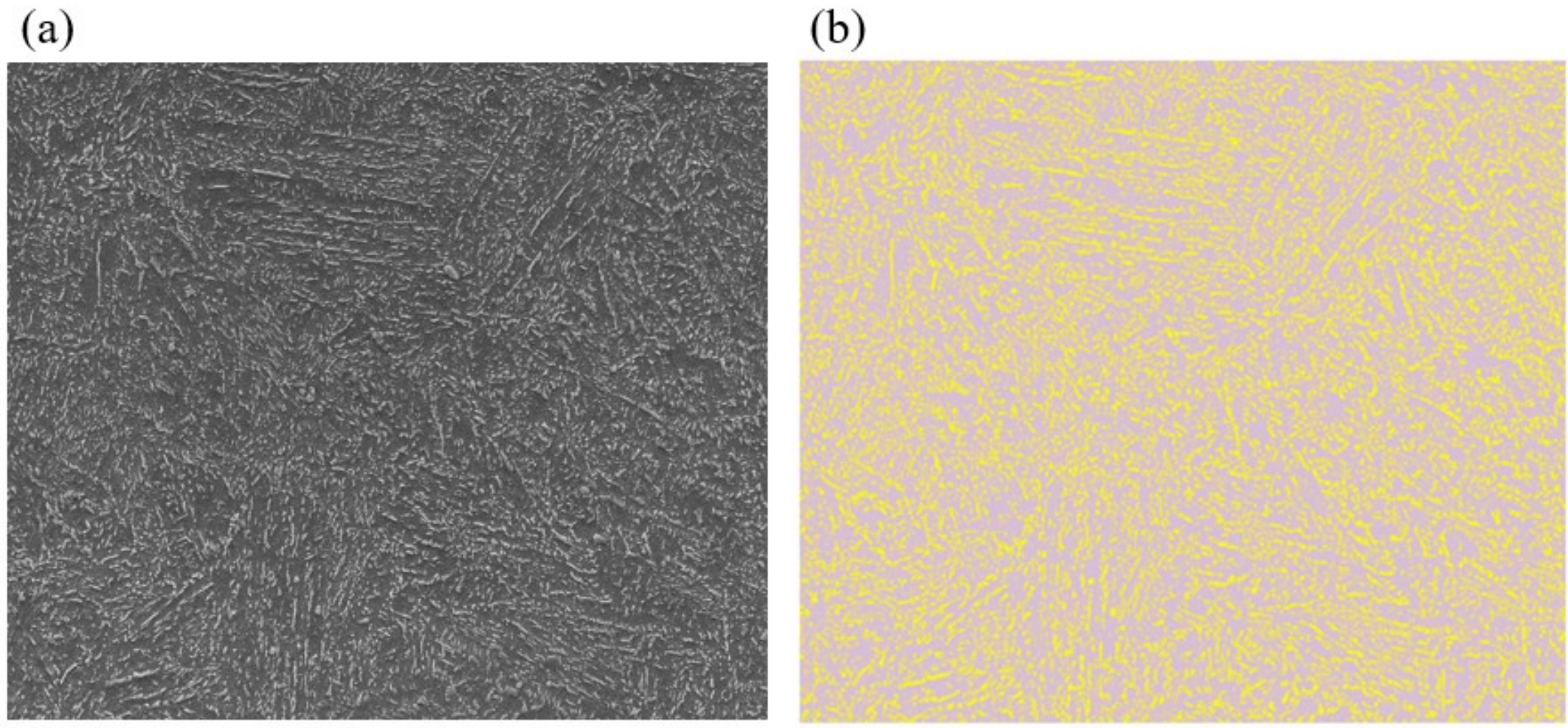

**Figure S7.** (a) Original uncropped TM SEM micrograph. (b) Reconstructed classifications showing the distribution of carbides (yellow) and the steel matrix (pink).

## S13. Calculation of Carbide Dimensions

Carbide sizes were derived from the merged segmentation masks (**see Figure S7b**) through a calibrated pixel-to-area conversion. First, the spatial resolution was determined by measuring the pixel density along the SEM scale bar (e.g., the 5 μm bar in **Figure S8**). This calibration factor yielded the physical area represented by a single pixel. Finally, the area of each individual carbide was computed by multiplying its total pixel count by the calibrated pixel area.

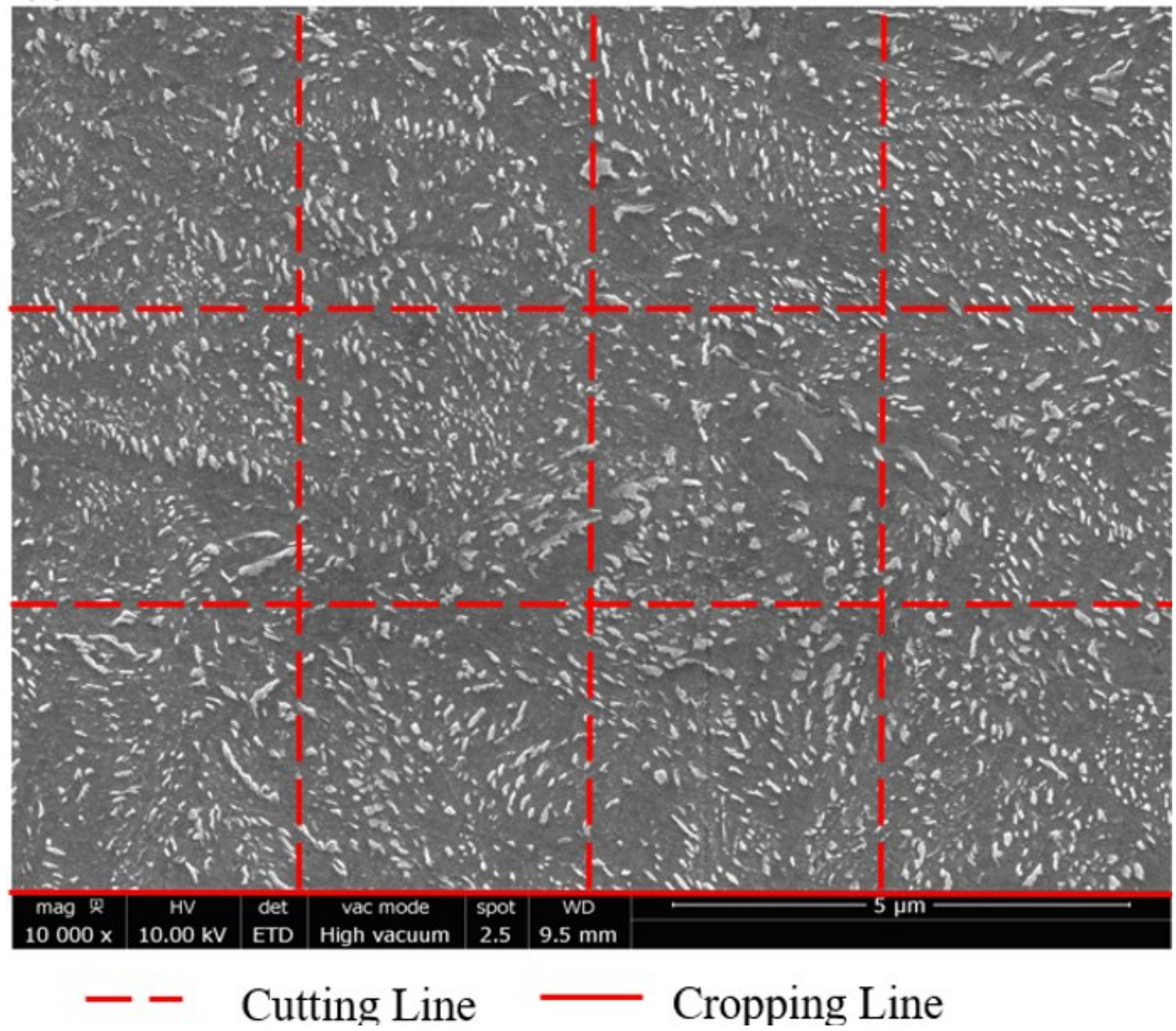

**Figure S7.** SEM image of the as-received LB. Scale bar represents 5 μm.